\PassOptionsToPackage{bookmarks=true,hidelinks}{hyperref}
\PassOptionsToPackage{sort&compress}{natbib}
\documentclass{article}

\usepackage[final]{corl_2026} 
\usepackage{times}
\usepackage{amsmath,amsfonts,amssymb}
\usepackage{notoccite}
\usepackage{amsfonts} 
\usepackage[ruled,vlined]{algorithm2e}
\usepackage{multicol}
\usepackage{xcolor}
\usepackage{subcaption}
\usepackage{multirow}
\usepackage[font=small,skip=2pt]{caption}
\usepackage{float}
\usepackage{wrapfig}
\usepackage{diagbox}
\usepackage{bm}
\usepackage{changepage}
\usepackage{booktabs}
\usepackage{multirow}
\usepackage{xcolor}  
\usepackage{svg}

\usepackage{outlines}

\usepackage{enumitem}
\setenumerate[2]{label=\alph*.}
\setenumerate[3]{label=\roman*.}

\newcounter{myenum}
\newenvironment{flushenumerate}{%
 \begin{list}{\arabic{myenum}.}%
   {\setlength{\leftmargin}{15pt}}%
    \setlength{\labelwidth}{20pt}
    \setlength{\itemindent}{0pt}
    \setlength{\labelsep}{0.5em}
 \setlength{\itemsep}{1pt}
 \setlength{\parskip}{0pt}
 \setlength{\parsep}{0pt}
    \usecounter{myenum}}%
{\end{list}}
\usepackage{color}

\usepackage[textsize=tiny]{todonotes}
\usepackage{ifthen}
\newboolean{comments}
\usepackage{multirow}
\setboolean{comments}{true} 
\ifthenelse{\boolean{comments}}{
  \setlength{\marginparwidth}{1.5cm}
  \newcommand{\harishnote}[1]{\todo[color=purple!20, size=\tiny]{Harish: #1}}
  \newcommand{\harishtext}[1]{\textcolor{purple}{Harish: #1}}
}{
  \renewcommand{\harishnote}[1]{}
  \renewcommand{\harishtext}[1]{}
}

\providecommand{\R}{\ensuremath \mathbb{R}}
\providecommand{\N}{\ensuremath \mathbb{N}}

\newcommand{\regtext}[1]{\mathrm{\textnormal{#1}}}

\newcommand{\vc}[1]{\mathrm{#1}}
\newcommand{\set}[1]{\mathcal{#1}}

\newcommand{\lossfunction}{\mathcal{L}}
\newcommand{\expectation}{\mathbb{E}}

\newcommand{\lbl}[1]{_{\regtext{#1}}}

\newcommand{\ubm}{\lbl{UBM}}
\newcommand{\enc}{\lbl{enc}}
\newcommand{\dec}{\lbl{dec}}

\newcommand{\dataset}{\set{D}}

\newcommand{\observation}{\vc{o}}
\newcommand{\action}{\vc{a}}
\newcommand{\latent}{\vc{z}}
\newcommand{\ubstate}{\vc{\xi}}
\newcommand{\image}{\vc{I}}
\newcommand{\liftedstate}{\vc{\psi}}
\newcommand{\visualfeatures}{\vc{\varphi}}

\newcommand{\UBM}{f\ubm}
\newcommand{\encoder}{f\enc}
\newcommand{\decoder}{f\dec}

\begin{document}



\title{Going with the Flow: Koopman Behavioral Models as Pseudo Planners for Visuo-Motor Dexterity}

\author{
Yunhai Han$^{1}$ \quad
Jiaqi Fu$^{1,\dagger}$ \quad
Linhao Bai$^{1,\dagger}$ \quad
Ziyu Xiao$^{1,\dagger}$ \\
\textbf{Zhaodong Yang}$^{1,\dagger}$ \quad
\textbf{Yogita Choudhary}$^{1,\dagger}$ \quad
\textbf{Krishna Jha}$^{1}$ \\
\textbf{Chuizheng Kong}$^{1}$ \quad
\textbf{Shreyas Kousik}$^{1}$ \quad
\textbf{Harish Ravichandar}$^{1}$ \\
$^{1}$Georgia Institute of Technology \\
$^{\dagger}$Equal contribution.
}

\maketitle

\begin{abstract}
Contemporary visuo-motor dexterity models often rely on expressive policy classes with diffusion and transformer backbones to achieve strong performance. 
However, these architectures require significant data and computational resources, and remain far from reliable, particularly for multi-fingered dexterity.
Importantly, they model skills as \textit{reactive mappings} and rely on fixed-horizon action chunking, creating a rigid trade-off between temporal coherence and reactivity.
To address these issues, we first introduce \textbf{Unified Behavioral Models (UBMs)}, a framework to represent dexterous skills as coupled dynamical systems that capture how visual features of the environment (\textit{visual flow}) and proprioceptive states of the robot (\textit{action flow}) co-evolve.
As such, UBMs ensure temporal coherence by construction rather than heuristic averaging. 
Unlike world models that attempt to predict the impact of \textit{arbitrary} robot actions on the environment, UBMs target \textit{behavioral dynamics} that encode how demonstrated robot behavior is related to  \textit{desired} impacts on the environment.
A UBM can be viewed as a \textit{pseudo planner}: given an initial condition, it computes the desired robot behavior over the entire skill horizon, while simultaneously ``imagining" the resulting flow of visual features.
We propose \textbf{Koopman-UBM} as a first instantiation of UBMs in the form of a \textit{structured latent linear} system.
K-UBM's inherent structure enables reactivity and adaptation via \textit{automated online replanning}: the model acts as its own runtime monitor, automatically triggering replanning when predicted and observed visual flow diverge beyond a threshold. Across seven simulated tasks and four real-world tasks, our approach performs comparably to state-of-the-art baselines, while offering considerably smoother executions, fast inference, efficient learning, robustness to occlusions, and automated online replanning.
\end{abstract}

\section{Introduction}


\begin{figure*}[t]
\centering
\includegraphics[width=\textwidth]{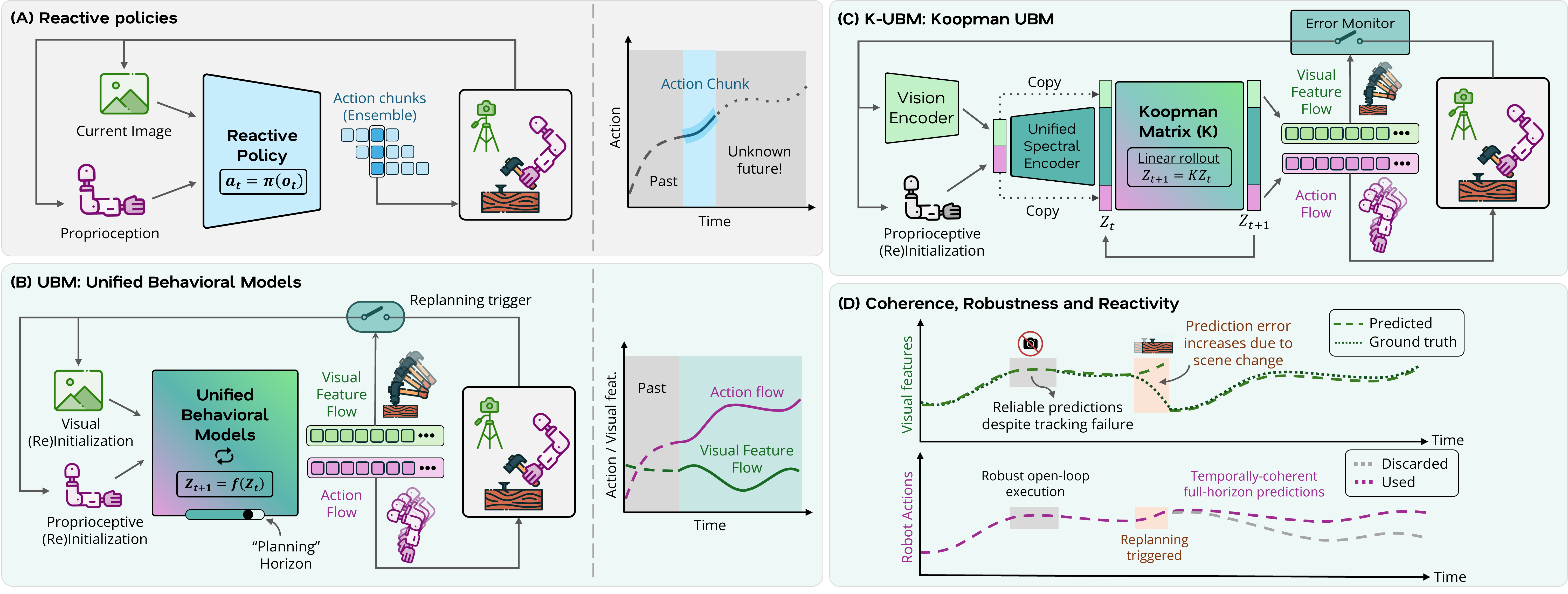}
\caption{
\textbf{(A)} Standard reactive policies (e.g., Diffusion, ACT) map observations to short-horizon action chunks, lacking a consistent internal model of the future or memory beyond the observation window, leading to temporal incoherence and hand-coded chunk lengths. 
\textbf{(B)} In contrast, \textit{Unified Behavioral Models (UBM)} model skills as joint behavioral dynamics of the robot and environment governing a continuous flow in a latent space, ensuring coherence and enabling flexible-horizon ``planning" from initial conditions. 
\textbf{(C)} We propose \textit{Koopman-UBM} as the first instantiation of UBMs, which lifts visual and proprioceptive observations into a latent space governed by a learned Koopman Operator.
By enforcing linear spectral dynamics ($z_{t+1} = Kz_t$) over a unified ``state-inclusive" latent space, K-UBM enables fast inference and predictive monitoring. 
\textbf{(D)} Our approach enables \textit{temporally-coherent} predictions (dashed-purple) that are \textit{robust} to visual occlusion (dashed green and purple trajectories inside gray boxes) and \textit{reactivity} via an event-triggered replanning strategy that reinitializes the UBM only when the predicted visual features diverge from reality (top orange box).
}
\label{fig:framework}
\end{figure*}

Reliable dexterous manipulation with multi-fingered hands has the potential to enable robots to seamlessly operate in a world made for and by humans.
Unlike simple pick-and-place with parallel-jaw grippers, multi-fingered dexterity requires mastering high-dimensional coordinated control under complex contact dynamics, partial observability, and frequent visual occlusion \cite{okamura2000overview, xu2025dexsingrasp, an2025dexterous, xu2025dexumi, yin2023rotating, yin2025dexteritygen, xu2025robopanoptes}. 
Data-driven approaches like Diffusion Policies \cite{reuss2023diffusion,diffusion_policy,ze20243d} and Transformers (e.g., ACT \cite{zhao2023learning}) have revolutionized this space by learning directly from visual demonstrations. Fundamentally, these approaches model skills as \textit{reactive mappings} ($o_t \to a_t$), treating the task as a sequence of independent decisions rather than a continuous physical process. 
To encourage temporal coherence and prevent jitter, these methods rely heavily on action chunking ($o_t \to a_{t:t+H}$) or temporal ensembling~\cite{zhao2023learning, liu2024bidirectional, liu2025immimic, arachchige2025sail} as heuristics that average predictions over fixed, hard-coded time windows.
While effective, these strategies introduce a rigid trade-off between deliberate planning (full-horizon predictions) and reactivity (short chunks), and any resulting coherence is a byproduct of averaging rather than a fundamental property of the learned representation (see Fig.~\ref{fig:framework}, Panel A).


An alternative perspective is to view any dexterous skill not as a sequence of decisions, but as being governed by an underlying \textit{dynamical system}. Seminal work on Dynamic Movement Primitives (DMPs)~\cite{ijspeert2013dynamical, saveriano2023dynamic} and numerous dynamical systems-based approaches~\cite{ravichandar2020recent, khansari-zadeh_2011_LearningStableNonlinear, bahl2020neural, van2024geometric, xie_2022_NeuralGeometricFabrics, guo2024surprising, li2025elastic} have long considered skills as solutions to fictitious dynamical systems. However, they tend to model the dynamics of the robot's movement alone, treating the environment as a static boundary condition. 
While highly effective for goal-reaching, this robot-centric view fails to capture the essence of dexterous manipulation: dexterity resides in the intricate coupling between the robot hand and the environment~\cite{okamura2000overview}. In dexterous tasks, the object's movement is equally if not more important than that of the robot. 

In this work, we introduce \textbf{Unified Behavioral Models (UBMs)}, a framework that learns to represent dexterous skills as \emph{coupled} dynamical systems directly from demonstrations (see Fig.~\ref{fig:framework}, Panel B). UBMs capture the robot-environment interdependence by learning the behavioral dynamics of the entire system that governs how visual features of the environment (\textit{visual flow}) and proprioceptive states of the robot (\textit{action flow}) co-evolve. Specifically, UBMs learn a unified latent space in which the joint state of the robot and object evolves according to a latent dynamical system. During inference, UBMs can be seen as ``\textit{pseudo planners}" that can generate complete ``plans" via open-loop latent rollout from any given initial condition, thereby promoting \textit{temporal coherence}.  Crucially, UBMs can trivially support \textit{dynamic chunking} as one can flexibly vary the number of time steps for which the model is rolled out (i.e., a flexible planning horizon).
Further, unlike increasingly-popular world models that attempt to predict the impact of \textit{arbitrary} robot actions on the environment, UBMs represent task-specific \textit{behavioral dynamics} that encode how demonstrated robot behavior is related to the  \textit{desired} impacts on the environment. Please refer to Appendix~\ref{sec:related_work} for a detailed discussion of related work.

Despite the many conceptual benefits, realizing UBMs presents a significant practical challenge: the joint dynamics of a multi-fingered hand interacting with a deformable or sliding object are highly nonlinear and difficult to learn directly from limited visual data without drifting or diverging. 

We introduce \textbf{Koopman-UBM} as the first instantiation of UBMs that leverages Koopman Operator theory~\cite{Koopman1931Koopman, mauroy2020koopman, abraham2017model, bruder2021koopman, shi2026koopman} to ensure tractable learning and efficient computation (see Fig.~\ref{fig:framework}, Panel C). By lifting the nonlinear interaction between the robot and visual features into a latent space where their joint evolution becomes linear ($z_{t+1} = Kz_t$), Koopman-UBM transforms the complex nonlinear dynamics learning problem into a representation learning problem for linear dynamics. 
After encoding the initial conditions into the latent space, Koopman-UBM can generate the entire nominal trajectory via linear rollout, ensuring temporal coherence by construction regardless of the rollout horizon. This allows Koopman-UBM to act as an implicit visuo-motor planner that predicts the future flow of the environment (e.g., object motion) alongside the robot’s actions without the computational cost of continuous sequential decision making via reactive mapping. 


To ground K-UBM in raw high-dimensional visual data, we fuse proprioceptive history with robust visual representations, and investigate the effectiveness of two complementary representations: (i) motion-centric features derived from point tracking (Object Flow~\cite{karaev2024cotracker}), and (ii) manipulation-centric features learned via self-supervised (DynaMo~\cite{cui2024dynamo}). In addition, we leverage a \textit{state-inclusive} lifting strategy~\cite{korda_linear_2018} (i.e., visual and proprioceptive features appear as a subvector of $z$) to form a latent state that explicitly embeds the system's structured state and enable fast decoder-free inference. By optimizing this representation jointly with the behavioral dynamics, our method learns \textit{dynamics-aware latent representations} that linearize complex contact behaviors. This helps ensure that Koopman-UBM's roll-outs are reliable and coherent during nominal execution. Unlike reactive policies that falter and freeze during strong occlusions and frame drops, K-UBM can faithfully propagate the system's ``momentum" without sensory feedback by relying on its internal dynamics model to bridge perceptual gaps (see Fig.~\ref{fig:framework}, Panel D).


However, relying solely on open-loop rollouts from initial conditions has a considerable limitation: we cannot account for unanticipated external disturbances during execution. To enable \textit{reactivity} without sacrificing coherence, we leverage UBM's unique ability to predict visual features to introduce a framework-level \textit{event-triggered replanning} strategy. Because Koopman-UBM can explicitly predict the nominal flow of visual features, the model can act as its own runtime monitor (see Fig.~\ref{fig:framework}, Panel D). Our replanning strategy monitors the error between predicted and observed visual features. The system executes the initial nominal plan as long as predictions match reality; when a disturbance causes the error to exceed a threshold, the system triggers a replan (re-initializes $z_{t}$ and computes a new coherent trajectory). Combining this flexible replanning strategy with Koopman-UBM allows the robot to flexibly navigate the spectrum between deliberate planning and reactive decision making without enforcing a rigid trade-off.


In summary, our key contributions are:
\begin{flushenumerate}
    \item \textbf{Unified Behavioral Models (UBM)}, a new class of sensory-motor skill models that encode dexterous skills as coupled dynamical systems, capturing the inter-dependence of robot actions and environmental features. UBMs offer temporal coherence by construction and enable full-horizon predictions of unified nominal actions and visual feature flows.
    
    \item \textbf{Koopman-UBM}, a Koopman-based instantiation of UBM that acts as an implicit visuo-motor planner. By leveraging state-inclusive lifting and linear spectral dynamics, Koopman-UBM enables reliable open-loop trajectory generation while also supporting closed-loop reactivity via predictive monitoring.
    
\end{flushenumerate}

Our \textbf{comprehensive evaluations} across seven simulated and four real-world dexterous manipulation tasks demonstrates that K-UBM performs comparably to state-of-the-art policy classes (e.g., Diffusion policy, ACT, etc.), while offering considerably smoother executions, fast inference, efficient learning, robustness to occlusions, and automated online replanning.

\section{Unified Behavioral Models}
\label{sec:UBM}

We first introduce the general problem of learning visuo-motor dexterous skills from demonstrations, and then introduce our unified behavioral models (UBMs) based solution.

\textbf{Problem Setup.}
Let $\dataset = [\{{\observation}_t^{(1)}, {\action}_t^{(1)}\}^{\bar{T}^{(1)}}_{t=1}, \cdots, \{{\observation}_t^{(N)},{\action}_t^{(N)}\}^{\bar{T}^{(N)}}_{t=1}]$ denotes a dataset of $N$ demonstrations of a certain target skill (e.g., opening a box, hammering a nail, etc.), with the $i$-th demo being a trajectory with $\bar{T}^{(i)}$ steps.
At each time step \(t\), the observation \(\observation_t\) consists of the robot joint state and visual observations, while the action \(\action_t\) corresponds to the robot joint command.
Specifically, we define the observation as $\observation_t = \{\mathrm{q}_t, \image_t\}$, where \(\mathrm{q}_t \in \mathcal{Q} \subseteq \R^{d_q}\) denotes the robot joint state (i.e., joint configuration of the arm and the hand) with $d_q$ degrees-of-freedom (dof), and \(\image_t \in \mathcal{I} \subseteq \R^{H \times W \times 3}\) denotes the RGB visual observation. 
Actions are defined as $\action_t = \mathrm{q}^{\mathrm{target}}_t \in \mathcal{Q} \subseteq \R^{d_q}$, representing the robot commands issued to the low-level controller. 

\textbf{Problem Statement.}
Design and learn a model $f$ from $\dataset$ to infer future actions based on current and past observations: $\hat{\action}_{t:t+t\lbl{p}}=f(\observation_{t-t\lbl{h}:t})$, where $t\lbl{p} \in \N$ and $t\lbl{h} \in \N$ denote the \textit{planning horizon} and the length of \textit{observation history}, respectively.
Note, this formulation respects causality: at each time step, the model is neither allowed leverage future observations nor alter past decisions.


\textbf{Contemporary Solution: Reactive Policies.} The most common approach to learning the dexterous skill encoded in $\dataset$ involves learning a reactive policy~\cite{diffusion_policy,zhao2023learning}. A reactive policy can be represented as a map 
$\pi:\ \observation_t\mapsto \action_{t}$. This policy can be generalized as $\observation_{t-H:t} \mapsto \action_{t:t+C}$, where $H\in\N$ and $C\in\N$ denote \textit{hard-coded, fixed} lengths of observation history and action chunks, respectively.

\textbf{Our Solution: Unified Behavioral Models.}
Our key idea is to represent the target dexterous skill as a dynamical system that encodes \textit{behavioral dynamics}: laws that govern how \textit{desired} visual features of the environment (visual flow) and proprioceptive states of the robot
(action flow) co-evolve.
Formally, we define a UBM as a map $\latent_{t+1} = \UBM(\latent_t)$, where $\latent_t=\encoder(\ubstate_t)$ is the latent state of the UBM produced by encoding the \textit{unified behavioral state} $\ubstate_t=[\action_{t},\observation_{t}]$, containing both action and observation information.
To extract the predicted actions from the latent flow, UBMs also require an action decoder: $\action_t=\decoder(\latent_t)$. As such learning a UBM requires learning $\UBM$ (the latent behavioral dynamics model), $\encoder$ (unified sensory-motor encoder), and $\decoder$ (action decoder) simultaneously.
While it might be possible to learn these modules separately, \textit{co-training} them is necessary to learn a temporally-coherent and predictive latent representation over which we could learn dynamics~\cite{assran_v-jepa_2025,garrido_learning_2026}.

Although closely related, note that UBMs are different from the increasingly popular notion of ``robotic world models"~\cite{ai_review_2025,li_robotic_2025}.
World models learn to predict the impact of a given \textit{arbitrary} robot action by approximating the underlying natural dynamics of the system (i.e., $\vc{x}_{t+1} = f\lbl{WM}(\vc{x}_t,\action_t)$ where $\vc{x}_t$ is the (latent) state of interest and actions come from, e.g., play data), whereas UBMs capture the \textit{desired} sensory-motor behavioral dynamics that underlie demonstrations.



UBMs offer a number of practical benefits when learning dexterous skills.
First, UBMs serve as pseudo planners that can generate ``plans" of arbitrary lengths up to the full skill horizon. 
Specifically, given an initial condition $\ubstate_0 = (\action_{0},\observation_{0})$, we can initialize the unified latent state $\latent_0=\encoder(\ubstate_0)$ and forward propagate (rollout) the learned dynamics  $\UBM(\cdot)$ up to \textit{any} desired planning horizon $t_p\leq \bar{T}$ (with $\bar{T}\in \N$ representing the full skill-horizon): $\{\latent_t\}_{t=1}^{t_p}=f_{UBM}(\latent_0)$. Finally, we can leverage the learned action decoder $\decoder(\cdot)$ to extract the corresponding action trajectory $\{\action_t\}_{t=1}^{t_p}$.


\section{K-UBM: Koopman-based Unified Behavioral Models}
\label{sec:spectral_policies}


Numerically creating a UBM is challenging, because it requires handling high-dimensional raw visual observations and action spaces, noisy data collection, non-smooth contact dynamics, and temporal drift in latent dynamical systems.
To address these challenges, we introduce \textit{Koopman-UBM}, a practical instantiation of UBMs using Koopman operator theory and motion-centric visual features; the key idea of Koopman operators is to learn a high-dimensional but \textit{linear} dynamical system representation.

Koopman-UBM combines representation learning and latent behavioral dynamics learning, and learns from demonstrations in two stages.
First, we perform \textit{Visual Feature Extraction} from raw images $\{\image_t\}$ across all time steps of the demonstrations (Section.~\ref{sec: visual_feature_extraction}).
Second, we learn a Koopman-based UBM to predict both visual feature and action flows given initial conditions (Section.~\ref{sec: koopman_learning}). 

\subsection{Extracting Visual Features}
\label{sec: visual_feature_extraction}
We begin by extracting and compressing relevant visual features from RGB images.
Without this crucial compression step, UBMs would attempt to learn the dynamics of task-irrelevant pixel-level information and potentially find spurious correlations or struggle to converge.
We investigate two visual representations. 
See Appendix.~\ref{Appendix:visual_feature_learning} for implementation details.
\newline
\textbf{Object Flow Points}: We track 256 points using SAM3 and CoTracker, then compress their coordinates into a 128-dimensional latent space using a convolutional autoencoder to provide a compact, motion-centric representation.
\newline
\textbf{DynaMo:} We utilize a ResNet-18 encoder trained with a self-supervised, dynamics-consistency objective that learns compact, dynamics-aware embeddings from task-specific RGB observations.
\subsection{Koopman Unified Behavioral Model Learning}
\label{sec: koopman_learning}
We learn Koopman UBMs from demonstrations by jointly training
i) a unified spectral encoder and ii) an approximated Koopman operator governing the linear evolution in unified latent space.

\textbf{Unified Behavioral State}:
We define the unified behavioral state by concatenating the robot actions
\(\action_t\) and learned visual features \(\visualfeatures_t\):
$
    \ubstate_t \triangleq
    [\action_t^{\ T},\visualfeatures_t^{\ T}]^{\ T}
    \in \R^{d_{\ubstate}},
$
where $d_{\ubstate} = d_q + d_{\visualfeatures}$ is the size of the unified behavioral state. To balance the magnitudes of the state components, we rescale the visual features $\visualfeatures_t$ by a constant factor computed based on the average $l_2$ norms of the visual features and the robot actions over all demonstrations.

\textbf{Unified Spectral Latent:}
The key idea is to approximate a Koopman-invariant subspace.
To do so, we first define a unified spectral encoder \(g_\theta:\ubstate_t \mapsto \liftedstate_t\) that acts as a lifting function, with $\theta$ as its learnable parameters.
We parametrize this encoder using a multi-layer perceptron (MLP) that computes the lifted behavioral state $\liftedstate_t \in \R^{d_\liftedstate}$ from the unified behavioral state $\ubstate_t$.
Then, we define a \textit{state-inclusive} latent representation of the lifted behavioral state as
$
    \latent_t
    \triangleq
    [\ubstate_t^{\ T},\liftedstate_t^{\ T}]^{\ T}
    \in \R^{d_\latent},
$
where $d_z=d_{\ubstate} + d_{\liftedstate}$ denotes the dimension of the Koopman-UBM's latent space. By explicitly including $\ubstate_t$ in the latent representation, we enable the latent Koopman dynamics to be grounded in the behavioral states, avoiding drifts and encouraging temporal coherence. Further, state-inclusive latent representations enable fast decoder-free inference.

\textbf{Linear Latent Dynamics:}
With the latent space established, we define a learnable Koopman matrix \(K\in\R^{d_z \times d_z}\) such that the dynamics in the latent space can be approximated as $\latent_{t+1} = K\,\latent_t$.

\textbf{Co-training the Encoder and Dynamics:}
To ensure that the learned latent representation of the unified behavioral state is dynamics-aware and will permit a linear structure, we co-train the linear dynamics jointly with the unified spectral encoder. We minimize a multi-step latent prediction loss over a prediction horizon \(H\) to encourage temporal coherence in the latent space and minimize drift. We compute multi-step latent predictions by repeatedly applying the Koopman operator on a given latent state:
$\hat{\latent}_{t+l} = K^l\,\latent_t$ for $l = 1,\ldots,H$.
We define our linear coherence loss as the prediction error against the ground-truth latent targets:
$\lossfunction\lbl{K-coherence}
    =
    \expectation_{t}\left[
    \sum_{l=1}^{H}
    \left\|
    K^l\,\latent_t
    -
    \latent_{t+l}
    \right\|_2^2
    \right]$.

Based on comprehensive experimentation (see Appendix~\ref{sec:Recipes}), we identified two key factors that influenced training stability:

\textbf{Learning rates and gradient clipping:} We found that training stability improves significantly when the Koopman matrix is updated at a slower rate than the unified spectral encoder. We believe this is because the Koopman matrix is applied repeatedly during multi-step prediction, and its gradients are more prone to explosion. As such, we recommend using a smaller learning rate for the Koopman matrix, along with gradient clipping.

\textbf{Identity initialization of the Koopman matrix:} We found that initializing the Koopman matrix \(K\) to the identity matrix is crucial for stability, especially in the early stages of training. Intuitively, an identity initialization corresponds to a \textit{static} latent space and updating the matrix (especially at a slower rate as noted above) allows the latent dynamics to gradually grow in a rapidly-evolving latent space (as the encoder learns at a faster rate).
\section{Evaluations in Simulation}
\subsection{Experimental Design}\label{sec:exp_design}
\begin{figure*}[t]
\centering
\includegraphics[width=1.0\textwidth]{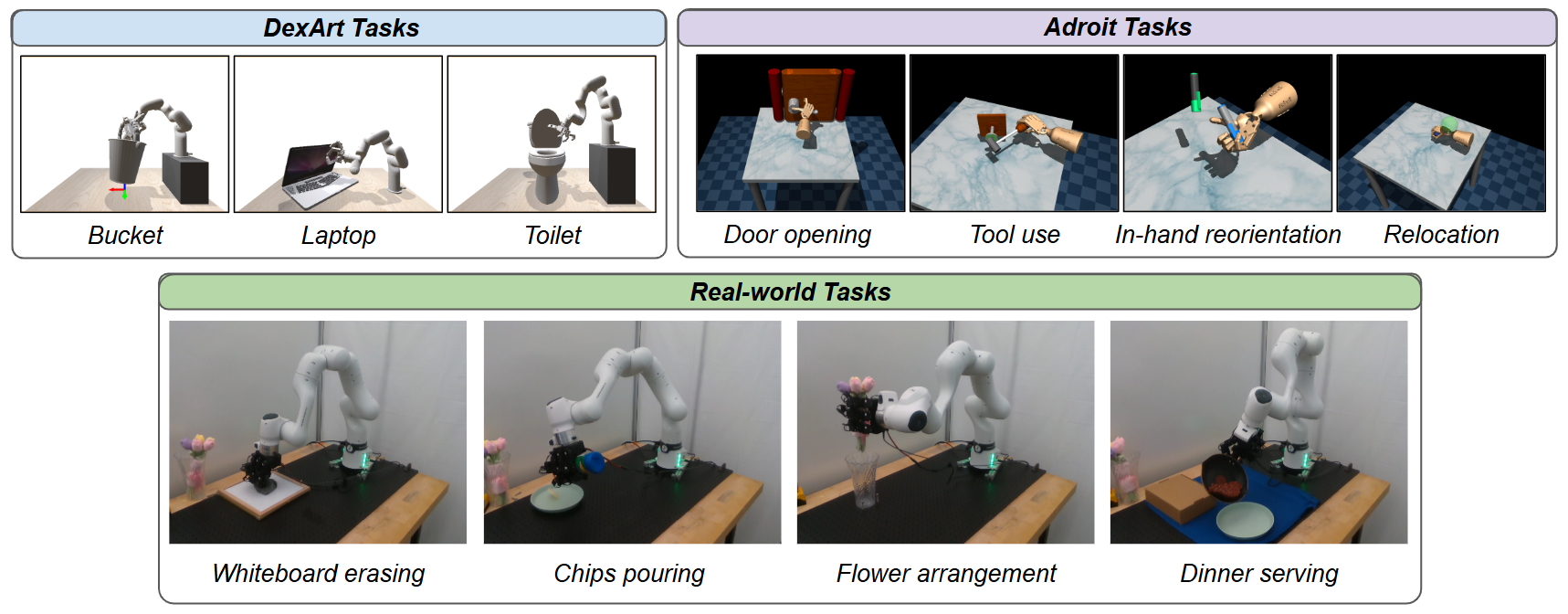}
\caption{We evaluate K-UBM on seven simulation tasks and four real-world tasks.}
\label{fig:tasks}
\end{figure*}

\noindent\textbf{Tasks}: We consider seven challenging and representative simulated tasks (See Fig.~\ref{fig:tasks}). 

\textit{DexArt}~\cite{bao2023dexart}.
In the SAPIEN simulator~\cite{xiang2020sapien}, the robot platform consists of a 6-DoF XArm6 arm and a 16-DoF Allegro hand, and is required to perform three challenging articulated object dexterous manipulation tasks: \textit{Bucket, Laptop, and Toilet}.

\textit{Adroit}~\cite{rajeswaran2017learning}.
In the MuJuCo simulator~\cite{Todorov2012MUJuCO}, the robot platform consists of an Adroit hand—a 30-DoF system comprising a 24-DoF hand and a 6-DoF floating wrist base, and is required to perform four representative dexterous manipulation tasks: \textit{Door opening, Tool use, In-hand reorientation, and Object relocation}.
Note that \textit{Tool use}, \textit{In-hand reorientation}, and \textit{Object relocation} are goal-conditioned tasks. For tasks involving explicit goals (e.g., moving an object to a target pose), we concatenate the current visual feature at each step with a fixed goal feature.

\textbf{Demonstrations}: 
For \textit{DexArt} tasks, we collect 30 demonstrations for training and 30 for testing. For \textit{Adroit} tasks, we collect 50 demonstrations for training and 30 for testing. For both simulation benchmarks, we use the pre-trained RL experts. Each demonstration consists of visual observations, robot states, and commanded actions. 

\textbf{Baselines}: We consider the following five policy architectures or classes as baselines, representing both contemporary standards and SOTA Koopman-based methods, including: i) Diffusion policy~\cite{diffusion_policy}, ii) Action Chunking Transformer (ACT)~\cite{zhao2023learning}, iii) UVA~\cite{li2025uva}: iv) KODex~\cite{han2023utility}, v) KOROL~\cite{chen_korol_2024}. See Appendix.~\ref{appendix:policy_designs} for detailed descriptions.

To ensure a fair comparison, we:
(i) designed the robot state and visual features for all methods to be identical,
(ii) designed the baseline policies and tuned their hyper-parameters for each baseline method (Appendices~\ref{appendix:policy_designs} and \ref{appendix:layer_size}), and
(iii) trained each policy over five random seeds to control for initialization effects~\cite{henderson2018deep}, except for \textit{KODex}, which computes an analytical solution.
\subsection{Comparisons against baselines}
\label{sec:general_eval}
\begin{figure*}[t]
\centering
\includegraphics[width=\textwidth]{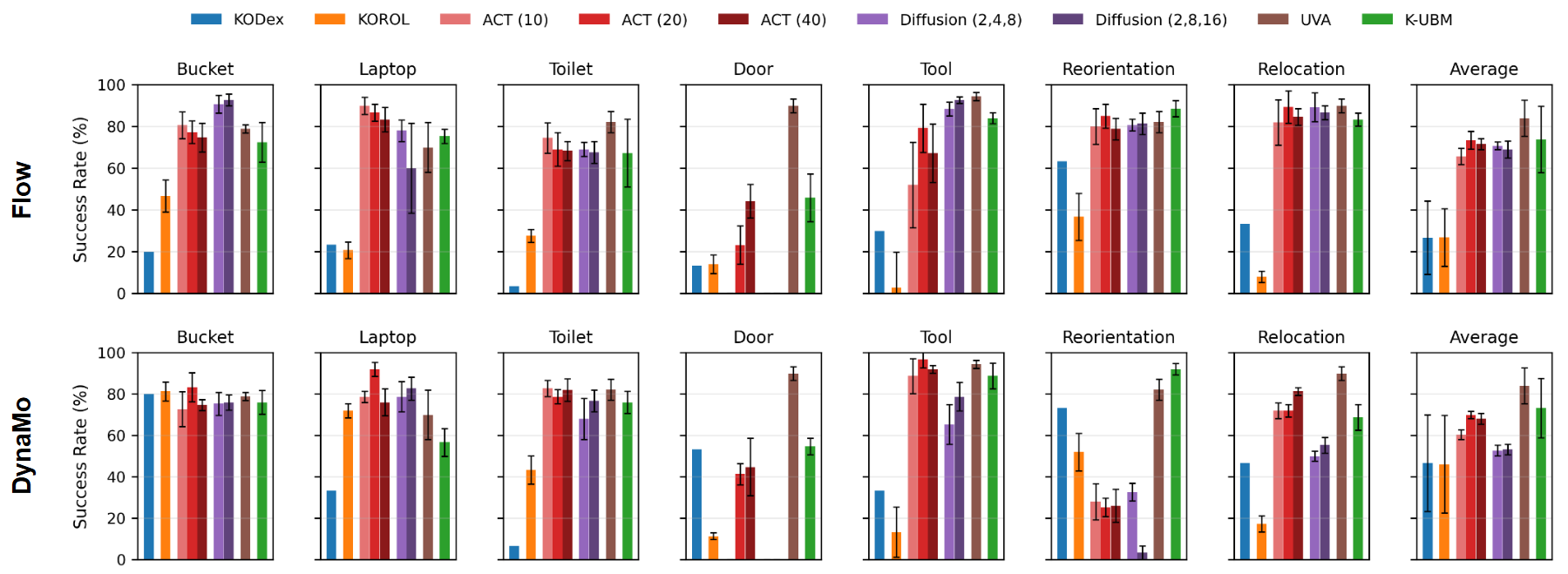}
\caption{We report success rates across all simulation tasks and both Flow and DynaMo features. Error bars indicate the standard deviation over five random seeds, and average results (last column) are computed over all seeds and all tasks. Since UVA operates directly on raw images rather than the precomputed visual features, its results are identical in both rows. ACT (10, 20, 40) denotes three ACT variants with different chunk lengths, while Diffusion (2, 4, 8) and (2, 8, 16) denote different execution and prediction horizons.}
\label{fig:simulation_results}
\end{figure*}

In this section, we discuss overall performance on all seven simulated tasks (across two types of visual features) in terms of task success rate and inference cost. 
In Fig.~\ref{fig:simulation_results}, we report task success rates on unseen test sets. On average, our K-UBM method achieves the second highest mean success rate. Compared to baselines, we observe that diffusion- and ACT-based policies require task-specific tuning of the action chunk size, highlighting the practical challenges of architecture selection for these models. In contrast, K-UBM achieves the most consistent performance without much architecture tuning. Compared to UVA, K-UBM achieves slightly lower performance but substantially faster inference, as shown in Table~\ref{tab:inference_cost}. Meanwhile, the inferior performance and high variance of KODex and KOROL may stem from the limitations of least-squares optimization, which is sensitive to noise, particularly when the input states include high-dimensional visual latents. Consequently, long-horizon predictions at inference become less robust, leading to poor task performance. By contrast, K-UBM adopts an end-to-end approach that jointly updates both the encoder and the Koopman matrix via gradient descent. 

We also evaluate the inference cost of each method for the bucket task\footnote{The inference cost is similar across tasks, so we report results on a single representative task for brevity.}. We report the inference cost in policy query time, which measures only the time required for policy inference, i.e., the total time the policy spends generating trajectories for task execution. The timing result is averaged over the total number of time steps, as different policies complete the task at different horizons. 

\begin{wraptable}{r}{0.5\columnwidth}
\centering
\small
\begin{tabular}{l cc}
\toprule
\textbf{Method} 
& \multicolumn{2}{c}{\textbf{Time (ms, $\downarrow$)}} \\
\cmidrule(lr){2-3}
& \textbf{Flow} & \textbf{DynaMo} \\
\midrule
KODex            & 0.40($\pm$0.04) & 0.37($\pm$0.04) \\
KOROL            & 0.41($\pm$0.07) & 0.36($\pm$0.04) \\
ACT              & 0.52($\pm$0.06) & 0.29($\pm$0.02) \\
Diffusion        & 29.95($\pm$1.4) & 36.31($\pm$2.3) \\
UVA$^\dagger$    & \multicolumn{2}{c}{98.6($\pm$26.9)} \\
K-UBM            & 0.42($\pm$0.05) & 0.47($\pm$0.14) \\
\bottomrule
\end{tabular}
\caption{Per-step inference (ms), excl. visual feature extraction.$^\dagger$UVA uses raw images.}
\label{tab:inference_cost}
\vspace{-1cm}
\end{wraptable}

From the results in Table.~\ref{tab:inference_cost}, we can observe that K-UBMs are highly efficient during inference because their action estimation relies on linear system rollouts, and can remain decoupled from online visual processing (required for initialization and event monitoring). The even monitor interrupts action estimation only when replanning is automatically triggered.

\subsection{Architecture Ablations}
\label{sec:architecture_ablation}
\begin{figure*}[t]
\centering
\includegraphics[width=\textwidth]{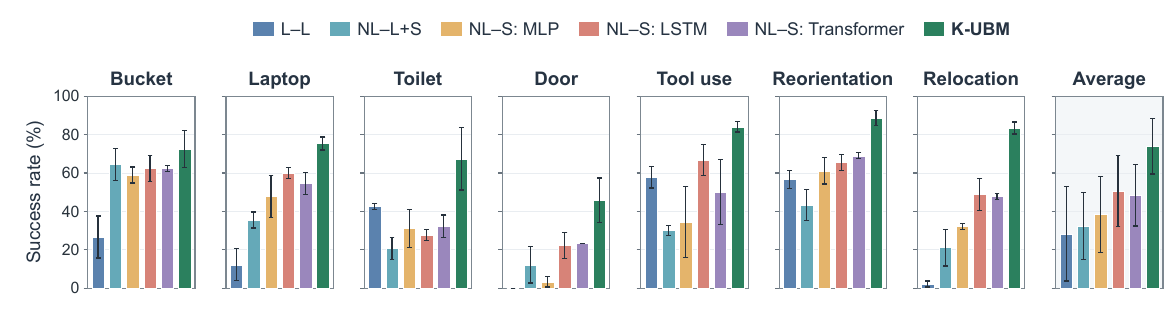}
\caption{Success rates of UBM architecture ablations. We report the mean and standard deviation across seven tasks (flow features) and three random seeds. K-UBM consistently outperforms the other architectures, highlighting the benefits of its two foundational design choices: state-inclusive lifting and linear latent dynamics.}
\label{fig:architecture_ablation}
\end{figure*}

We evaluate two foundational design choices of K-UBM: state-inclusive lifting and linear latent dynamics. Specifically, we compare K-UBM against linear latent-only dynamics (L-L), which exclude the original state; nonlinear latent-plus-state dynamics (NL-L+S), which replace linear dynamics with a nonlinear MLP; and nonlinear state dynamics (NL-S), which directly predict future states from current states using MLPs, LSTMs, or Transformers. As shown in Fig.~\ref{fig:architecture_ablation}, across seven tasks and three seeds with flow features, K-UBM achieves the highest mean success rate among all tested variants, demonstrating the effectiveness of our two design choices in learning effective behavioral dynamics. 

\vspace{-0.2cm}
\subsection{Additional Experiments}
\label{sec:additional}
\vspace{-0.2cm}
In Appendix.~\ref{appendix:visual_fea_pred}, we evaluate the visual feature prediction. In Appendix.~\ref{appendix:Reactivity}, we evaluate the policy reactivity enabled by the online replanning. In Appendix.~\ref{sec:Perturbation}, we conduct systematic experiments evaluating the robustness of reactive policies to camera noise, highlighting the benefits of reliable prediction under tracking failures. Additionally, we report ablation experiments on the Koopman learning recipe in Appendix.~\ref{sec:Recipes}.

\section{Real-world Evaluation}
\begin{figure*}[t]
\centering
\includegraphics[width=0.99\textwidth]{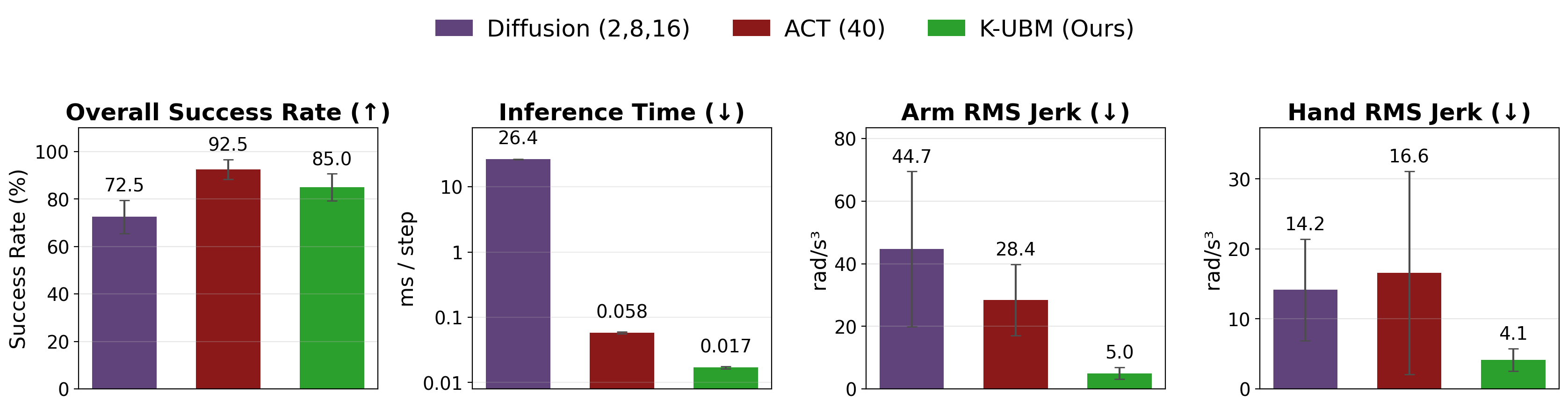}
\caption{Real-world performance across four tasks. K-UBM achieves comparable success rates with the fastest inference and lowest arm/hand jerk. Error bars denote standard deviation across task-level means.}
\label{fig:real_world_bar}
\end{figure*}

\begin{figure*}[t]
\centering
\includegraphics[width=0.95\textwidth]{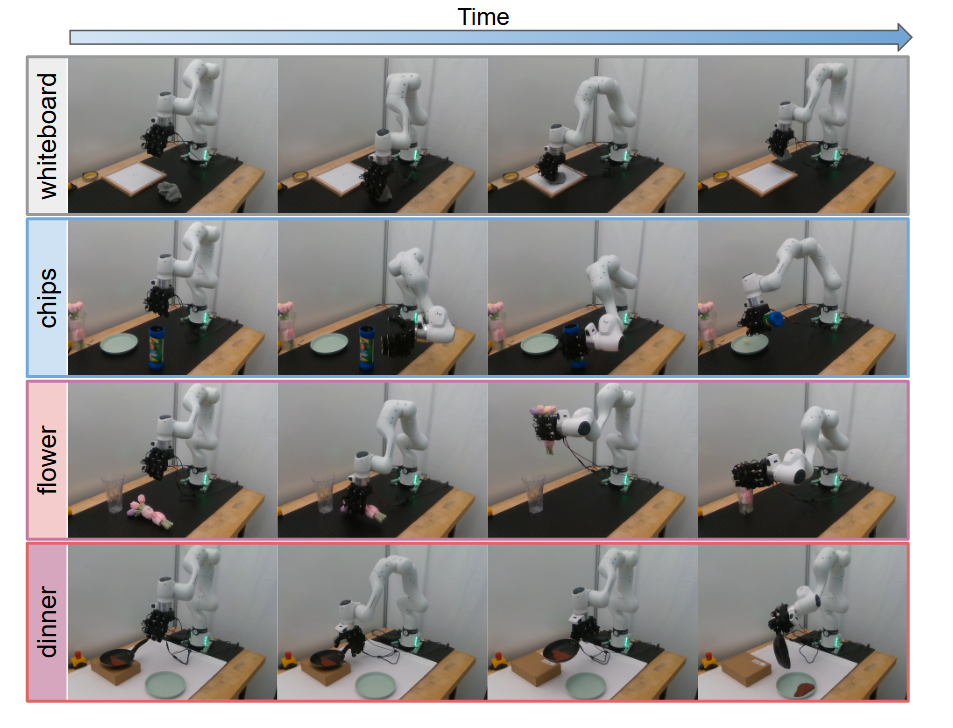}
\caption{Real-world K-UBM rollouts on the four manipulation tasks (top to bottom:
whiteboard erasing, chips pouring, flower arrangement, dinner serving). Each row
shows four time steps of one rollout.}
\label{fig:real_world_tasks}
\end{figure*}
In this section, we demonstrate that K-UBM can also be applied to real-world dexterous manipulation. The robot platform consists of a 7-DoF Franka arm paired with a 16-DoF LEAP hand~\cite{shaw2023leaphand}, yielding a 23-dimensional state space. We consider four manipulation tasks: i) Flower Arrangement, ii) Chips Pouring, iii) Whiteboard Erasing, and iv) Dinner Serving. See Fig.~\ref{fig:tasks} for snapshots.


For each task, we collect 20 demonstrations via kinesthetic teaching~\cite{akgun2012trajectories}, each consisting of synchronized visual observations, robot states, and commanded actions. We compare K-UBM against two strong baselines, Diffusion and ACT. We omit UVA due to its prohibitively long training time.



In Fig.~\ref{fig:real_world_bar}, we present a concise across-task summary of success rate, per-step inference time, and trajectory smoothness (arm and hand jerk) for all methods across the four tasks (check Table.~\ref{tab:realworld_pertask} in Appendix.~\ref{appendix:real-world-per-task} for per-task results). While ACT attains the highest average success rate, K-UBM remains competitive (85.0\% vs.\ 92.5\%). More importantly, K-UBM produces substantially smoother motion, reducing arm jerk by roughly $6$--$9\times$ and hand jerk by $3$--$4\times$ relative to the Diffusion and ACT baselines, due to its dynamical-systems formulation that produces naturally coherent trajectories. It is also the fastest at inference, requiring only $0.017$~ms per step, further highlighting its efficiency from linear rollout and decoupling from visual feedback. These results show that K-UBM not only performs comparably to strong baselines, but also offers markedly more coherent motion at negligible inference cost. Additionally, see Fig.~\ref{fig:real_world_tasks} for K-UBM rollout snapshots for each task.

\section{Conclusions, Limitations, and Future Work} 
\label{sec:conclusion}
We introduced Unified Behavioral Models (UBMs), a framework treating dexterous manipulation as the coupled evolution of robot and environment dynamics in a shared latent space. We proposed Koopman-UBM, with several crucial design choices to enable effective learning of visuo-motor behavioral dynamics, enabling both temporal coherence and fast inference. Across seven simulated and four real-world tasks, K-UBM bridges dynamical systems theory with modern visual learning, matching SOTA baselines while offering superior trajectory coherence, low inference cost, occlusion robustness, and event-triggered replanning.
\vspace{+0.2cm}
\newline
\textbf{Limitations}: While promising, our current instantiation of UBMs has three primary limitations. First, while K-UBM's linear dynamics capture coarse-grained behavior, they smooth over sharp contact discontinuities; explicit hybrid modeling (e.g., switching Koopman operators) may be needed for high-frequency impact tasks~\cite{bakker2020learning}.
Second, our ``pseudo planning" via open-loop rollout might not be sufficient in highly stochastic environments with changing objects dynamics, and will likely require online adaptation or residual corrections~\cite{han2022desko}. Third, our reactivity relies on visual feedback; simultaneous occlusion and disturbance can delay replanning and cause failure, which may benefit from tactile feedback~\cite{yu2024mimictouch}.
\vspace{+0.2cm}
\newline
\textbf{Future Work:} The UBM framework opens several exciting directions for future research.
Future work could explore more expressive architectures for UBM (e.g., Neural ODEs~\cite{chen2018neural}, Transformers, Diffusion), and incorporate active perception for robust predictive monitoring and reward signals. Additionally, integrating UBMs into hierarchical control (in which foundation models, such as VLMs, modulate attractors or switch between primitives) could enable language-driven, long-horizon tasks. Ultimately, UBMs provide a promising alternative to reactive policies, flexibly blending forethought with reflex.

\clearpage
\bibliography{references}

\clearpage

\appendix

\section{Related work}
\label{sec:related_work}

Our work sits at the intersection of several sub areas: visuo-motor imitation learning, dexterous manipulation, dynamical-systems based skill models, Koopman-based robot learning, and the rapidly growing literature on robot world models and video-action models.. Below, we contextualize our contributions within these areas and position K-UBM as a compact, dynamics-centric alternative to reactive action predictors and large-scale video-action generative models.

\textbf{Visuo-Motor Imitation Learning and Action Chunking}
Modern visuo-motor imitation learning has made impressive progress by training expressive policies that map visual and proprioceptive histories directly to actions. Action Chunking Transformers (ACT)~\cite{zhao2023learning} and Diffusion Policies~\cite{diffusion_policy}. These methods have become strong baselines and serve as the default backbone for many modern policy and model architectures. Subsequent work has improved these policies through better spatial representations, such as point-cloud inputs in 3D Diffusion Policy~\cite{ze20243d}, closed-loop resampling mechanisms for action chunks~\cite{liu2024bidirectional}, flow-matching action generators such as $\pi_0$~\cite{black2024pi0}, and data-efficient variants for goal-conditioned or cross-domain imitation~\cite{reuss2023diffusion} Their success is often attributed to their ability to accurately predict actions given sufficient amount of high-dimensional and multi-modal data. 

Despite their strong performance on a variety of tasks, these models still represent a skill as a reactive map from recent observations to a finite action sequence (i.e., $ o_{t-h:t} \mapsto a_{t:t+H}$), where the temporal structure of the behavior is mediated by the chosen history length, prediction horizon, execution horizon, and any temporal averaging or ensembling. While powerful, this design inherently forces a trade-off between temporal coherence and reactivity. Further, these policies often only predict actions, ignoring predictive power over the robot's impact on its environment. In contrast, our goal is to learn a \emph{unified behavioral dynamics model}: a coupled dynamical system, defined over a structured latent space, whose rollout jointly predicts the co-evolution of desired robot actions and the environmental visual features.

\textbf{Dexterous Manipulation}

Dexterous manipulation is a particularly demanding setting for robot learning because success depends on high-dimensional coordination, intermittent contact, friction, deformation, partial observability, and tight coupling between the robot and the manipulated object~\cite{bicchi2000robotic,okamura2000overview, rajeswaran2017learning,bao2023dexart,an2025dexterous}. Recent work has made substantial progress through reinforcement learning, teleoperation, imitation, tactile sensing, cross-embodiment interfaces, and foundation-style dexterity controllers ~\cite{yin2023rotating,yu2024mimictouch,xu2025dexumi, yin2025dexteritygen,xu2025robopanoptes,xu2025dexsingrasp}. These efforts make clear that, in dexterous manipulation, the object trajectory and contact evolution are not merely consequences of the robot's motion; they are part of the skill itself.

\textbf{Robot Skills as Dynamical Systems}

A long line of structured skill-learning methods has sought to make robot learning more reliable by embedding useful inductive biases offered by dynamical systems into the policy representation. Dynamic Movement Primitives (DMPs) encode motor behaviors as stable attractor systems with useful goal and temporal modulation properties~\cite{ijspeert2013dynamical, saveriano2023dynamic}. Stable dynamical-systems approaches, including SEDS and related methods, learn continuous motion fields from demonstrations while enforcing convergence or stability constraints~\cite{khansari-zadeh_2011_LearningStableNonlinear}. Neural Dynamic Policies and Neural Geometric Fabrics similarly inject dynamical-system or geometric structure into high-dimensional visuo-motor policies~\cite{bahl2020neural, xie_2022_NeuralGeometricFabrics,van2024geometric}. More recent adaptive dynamical-system policies, such as Elastic Motion Policy, continue this theme by emphasizing robust one-shot adaptation through structured motion representations~\cite{li2025elastic}.

The common lesson from this literature is that structure can make learning more data-efficient, stable, and coherent. However, most of these approaches are fundamentally \textit{robot-centric}: they model joint-space, end-effector, or task-space motion, while the environment enters as a goal, constraint, or context variable. This is sufficient for many motion-reproduction problems, but it is limiting for dexterous manipulation, where the essential behavior is the coupled evolution of the hand, the object, the contact state, and the visual scene. K-UBM extends the dynamical-systems view from robot motion to \emph{unified behavioral dynamics}, making the manipulated environment an explicit part of the learned skill model.

\textbf{Koopman Operators for Robot Learning}
Koopman operator theory provides a principled way to represent nonlinear dynamics through linear evolution of lifted observables ~\cite{Koopman1931Koopman,mezic2020koopman,mauroy2020koopman}. Finite-dimensional approximations such as EDMD and related data-driven linear predictors have been used for nonlinear system identification and control~\cite{williams2015data,korda_linear_2018}, while deep Koopman methods learn lifting functions jointly with latent linear dynamics~\cite{lusch2018deep}. In robotics, Koopman methods have been explored for model-based control, soft robotics, hybrid systems, and data-driven policy learning because they combine nonlinear representation learning with the computational simplicity of linear rollout and control~\cite{abraham2017model,shi2021acd, bruder2021koopman,bakker2020learning,han2022desko,shi2026koopman}.

Koopman structure has recently proven especially useful for dexterous manipulation. KODex learns Koopman-based models of desired hand-object evolution and shows that linear rollout in a lifted space can yield sample-efficient and computationally efficient dexterous policies ~\cite{han2023utility}. KOROL extends this idea toward vision-based manipulation by learning visualizable object features that can be propagated through Koopman rollout, reducing dependence on ground-truth object state at runtime~\cite{chen_korol_2024}. These works demonstrate that Koopman operators can provide a useful inductive bias for contact-rich manipulation, especially when the goal is to learn a structured evolution model from limited data.

K-UBM builds on this perspective but changes the modeling target. Rather than using Koopman rollout only as a hand-object state predictor or object-feature propagator, we use it as a \emph{unified behavioral model} over action and visual feature flow. The state-inclusive lifting embeds the current behavioral state directly in the lifted representation, enabling decoder-free action extraction and reducing long-horizon latent drift. As a result, the same linear rollout produces both future actions and predicted future visual features. Importantly, visual prediction is not merely an auxiliary output of K-UBM, but the mechanism that enable event monitoring at runtime and automatic replanning.

\textbf{Robot World Models and Video Generation Models}

A rapidly growing line of work studies predictive models of the world as a basis for robot learning and planning. Classical latent world-model approaches such as PlaNet and Dreamer learn compact predictive dynamics from pixels and perform control by planning or policy improvement in imagination~\cite{hafner2019planet, hafner2023dreamerv3}. DayDreamer recently demonstrated that such world-model-based learning can be applied directly to physical robots across arms, legged robots, and mobile robots~\cite{wu2023daydreamer}. Recent robotic world-model work further studies neural simulators and learned dynamics models for manipulation, locomotion, partial observability, and robust policy optimization~\cite{li_robotic_2025,ai_review_2025}. These methods highlight a central idea that is also important for our work: predicting future environment evolution can provide useful structure for control. However, they are typically action-conditioned models of the world dynamics rather than models of the nominal behavior encoded in a demonstrated skill.

A related and increasingly popular direction uses image or video generation as a planning or policy interface. UniPi formulates policy learning as text-conditioned video generation, generating future frames and then extracting actions from the generated visual plan ~\cite{du2023unipi}. SuSIE uses pretrained image-editing models to propose visual subgoals that can be reached by a low-level controller ~\cite{black2023susie}. UniSim learns an interactive real-world simulator from heterogeneous data and uses generated experience for policy learning~\cite{yang2024unisim}. RoboDreamer learns compositional video world models for imagining future plans under novel object-action combinations~\cite{zhou2024robodreamer}. GR-1 and GR-2 use large-scale video generative pretraining to improve language-conditioned visual robot manipulation, predicting actions and future images within GPT-style or video-language-action architectures ~\cite{wu2024gr1,cheang2024gr2}. Gen2Act generates human videos in novel scenarios and uses the generated motion to improve robot manipulation~\cite{bharadhwaj2025gen2act}. Video Policy pushes this idea further by arguing that video generators can serve as robot policies when paired with action decoders~\cite{liang2025videopolicy}. Self-supervised predictive video models such as V-JEPA 2 also suggest that web-scale video prediction can provide physical representations that support robotic planning after limited robot-data adaptation ~\cite{assran_v-jepa_2025}. Latent-action world models extend this direction by learning action-like control variables from videos without explicit action labels~\cite{garrido_learning_2026}.

The closest recent developments are unified video action (UVA) and world action models (WAMs), which explicitly model future observations and actions jointly. UVA learns a joint video-action latent representation and uses decoupled lightweight diffusion heads for video and action decoding. Through masked training, the same model can serve as a policy, forward dynamics model, inverse dynamics model, and video generator~\cite{li2025uva}. Unified World Models (UWM) couple video and action diffusion within a unified transformer using independent diffusion timesteps for each modality, allowing a single model to represent policies, forward dynamics, inverse dynamics, and video prediction while leveraging both robot data and action-free video ~\cite{zhu2025uwm}. Concurrent work, such as DreamZero introduces World Action Models (WAMs) built on video diffusion backbones that jointly predict future world states and actions, enabling stronger physical-motion generalization than conventional VLAs in real-robot experiments ~\cite{ye2026dreamzero}. MotuBrain further scales the WAM paradigm with multimodal data, multi-view modeling, shared cross-embodiment action representations, and deployment-oriented inference optimizations~\cite{xiang2026motubrain}.

K-UBM is closely aligned with this broader movement toward predictive action-observation modeling, but occupies a different point in the design space. Video-generation-based world models and WAMs aim to predict high-dimensional visual futures, often to support broad generalization across tasks, instructions, and embodiments. K-UBM instead predicts compact manipulation-centric visual features together with robot actions for a particular dexterous skill family. This sacrifices the generality of large video generative models, but yields a lightweight behavioral dynamics model with efficient full-horizon rollout, explicit temporal coherence, and a direct prediction-error signal for runtime monitoring. Rather than generating photorealistic future video, K-UBM rolls out the task-relevant visual-action flow in a Koopman-structured latent space, making online inference highly efficient and directly useful for event-triggered replanning.
\section{Visual Feature Learning}
\label{Appendix:visual_feature_learning}
In this section, we describe our key design choices to extract the relevant visual features from images and compress them into a compact representation. Without this crucial step, UBMs would attempt to learn the dynamics of task-irrelevant pixel-level information and potentially find spurious correlations or struggle to converge. In this work, we investigate two visual representations: i) object flow point features~\cite{karaev2024cotracker}, and ii) DynaMo visual features~\cite{cui2024dynamo}.

\textbf{Object Flow Points}: Before extracting and tracking flow points, we first leverage SAM3~\cite{carion2025sam} with a language prompt describing the objects of interest to generate an object-centric segmentation mask. Based on this mask, we sample initial object flow points and track them over time using CoTracker~\cite{karaev2024cotracker}, yielding $N$ sequences of object flow points \(\{\mathrm{p}_t^{(i)}\}_{t=0}^{T^{(i)}}\) for \(i = 1, \ldots, N\). Each
\(\mathrm{p}_t^{(i)} \in \mathbb{R}^{256 \times 2}\) represents the image coordinates of 256 tracked pixel points at time \(t\). 

To obtain a compact representation suitable for Koopman-based dynamics learning, we train a convolutional autoencoder on the extracted object flow trajectories. 

\noindent \textit{Representation:}
At each time step \(t\) of the \(i\)-th demonstration, we represent the object flow as a set of 256 2D points
\(\mathrm{p}_t^{(i)} = \{(u_{t,j}^{(i)}, v_{t,j}^{(i)})\}_{j=1}^{256}\)
in image coordinates. We reshape these points into a spatial grid of size
\(16 \times 16\) and treat it as a two-channel image:
$\mathrm{x}_t^{(i)} \in \mathbb{R}^{2 \times 16 \times 16},
$
where the two channels correspond to the horizontal and vertical coordinates, respectively.

\noindent \textit{Encoder:}
The encoder \(\mathcal{E}\) consists of a stack of strided convolutional layers with stride \(2\),
which progressively downsample the spatial resolution while increasing the channel capacity.
Starting from an input resolution of \(16 \times 16\), the encoder applies a single strided convolution to downsample the spatial
resolution to \(8 \times 8\) while expanding the channel dimension from 2 to 8,
followed by a \(1\times1\) convolution that projects the features back to a
two-channel latent flow representation:
$
\mathrm{z}_t^{(i)} = \mathcal{E}(\mathrm{x}_t^{(i)}) \in \mathbb{R}^{2 \times 8 \times 8}
$, resulting in a 128-dimensional flow feature obtained by flattening $\mathrm{z}_t^{(i)}$.

\noindent \textit{Decoder:}
The decoder \(\mathcal{F}\) mirrors the encoder structure using transposed convolutions to upsample
the latent feature map back to the original spatial resolution, followed by a ReLU activation:
$
\hat{\mathrm{x}}_t^{(i)} = \mathcal{F}(\mathrm{z}_t^{(i)}) \in
\mathbb{R}^{2 \times 16 \times 16}.
$

\noindent \textit{Training Objective:}
We train the flow autoencoder by minimizing a reconstruction loss between the reconstructed and
ground-truth flow points:
\[
\mathcal{L}_{\text{flow}}
=
\mathbb{E}_{i,t}\!\left[
\left\| \hat{\mathbf{x}}_t^{(i)} - \mathbf{x}_t^{(i)} \right\|_2^2
\right].
\]
We aggregate all flow frames \(\mathbf{x}_t^{(i)}\) across demonstrations and treat each time step
independently during training. The model is optimized using Adam with an exponential learning-rate decay schedule.


\textbf{DynaMo:}
As an alternative to flow-based representations, we adopt \emph{DynaMo} visual features~\cite{cui2024dynamo}, which provide compact and dynamics-aware representations directly from raw RGB observations. DynaMo is a self-supervised, in-domain visual representation learning method that exploits the temporal structure of expert demonstrations by modeling latent visual dynamics, without requiring action annotations, data augmentations, or out-of-domain pretraining data.

\noindent \textit{Encoder:}
Given a set of expert demonstrations consisting of image sequences $\{o_1, \ldots, o_T\}$, DynaMo jointly trains a visual encoder and latent dynamics models directly on task-specific observations. Each RGB image $o_t \in \mathbb{R}^{3 \times 224 \times 224}$ is mapped to a latent embedding
\begin{equation}
    s_t = f_\theta(o_t) \in \mathbb{R}^d,
\end{equation}
where $f_\theta$ is a ResNet-18–based encoder. The encoder is pretrained entirely in-domain using only demonstration images from the target task.

\noindent \textit{Latent Dynamics Modeling:}
To capture the underlying visual dynamics, DynaMo introduces two auxiliary models operating in latent space: (i) a latent inverse dynamics model $q(z_{t:t+h-1} \mid s_{t:t+h})$, and (ii) a forward dynamics model $p(\hat{s}_{t+1:t+h} \mid s_{t:t+h-1}, z_{t:t+h-1})$. Here, $z_t \in \mathbb{R}^m$ represents a low-dimensional latent transition variable between consecutive embeddings, with $m \ll d$, preventing the latent from trivially encoding the next state. Both the inverse and forward dynamics models are implemented as causally masked Transformer encoders.

\noindent \textit{Training Objective:}
The encoder and dynamics models are trained end-to-end using a latent-space dynamics consistency objective. Specifically, the forward model predicts the next-step embeddings $\hat{s}_{t+1:t+h}$, which are compared against the target embeddings $s^{*}_{t+1:t+h}$ using a cosine similarity loss:
\begin{equation}
    \mathcal{L}_{\text{dyn}}(\hat{s}_t, s^*_t)
    =
    1 - \frac{\langle \hat{s}_t, s^*_t \rangle}
    {\|\hat{s}_t\|_2 \cdot \|s^*_t\|_2}.
\end{equation}
To avoid representational collapse, the target embedding is stop-gradient detached, i.e., $s^*_t = \text{sg}(s_t)$. In addition, a covariance regularization loss is applied over a minibatch of embeddings $S$ to encourage feature decorrelation:
\begin{equation}
    \mathcal{L}_{\text{cov}}(S)
    =
    \frac{1}{d}
    \sum_{i \neq j}
    \mathrm{Cov}(S)_{i,j}^2 .
\end{equation}
The final training objective is given by
\begin{equation}
\mathcal{L} = \mathcal{L}_{\text{dyn}} + \lambda \mathcal{L}_{\text{cov}}, \quad \lambda = 0.04 .
\end{equation}


\section{Task Descriptions}
\label{Appendix:task_setup}
In this section, we describe the task state space for both simulated and real-world tasks, along with the task objectives, sampling ranges, and success criteria for each.
\subsection{Dexart Benchmarks}
We follow the original state space design used in prior RL benchmark work for these three articulated object dexterous manipulation tasks~\cite{bao2023dexart} in Sapien simulator~\cite{xiang2020sapien}, where the robot proprioception space is the same across tasks and is defined by the 6-DoF XArm6 arm joint positions and velocities together with the 16-DoF Allegro hand finger joints (28 dimensions in total). We use the provided RL expert checkpoints to generate rollouts and collect successful trajectories for imitation learning, consisting of visual observations, observed robot states, and target robot commands. We also adopt the original sampling strategies during data collection and the success criteria for imitation learning policy evaluation, which are detailed below.

\textbf{Bucket}
This task requires the robot to lift a bucket. To ensure stable lifting, the robot must extend its hand beneath the bucket handle and grasp it to achieve form closure. This task introduces randomness in the bucket assets, including variations in size and geometry, as well as random initial object positions by adding independent uniform noise of up to 5 cm along each axis relative to the annotated positions. The task is considered successful if the bucket is lifted by at least 30 cm within a specified time duration.

\textbf{Laptop} This task requires the robot to grasp the middle of the screen and then open the laptop lid. This task introduces randomness in the laptop assets, including variations in size and geometry, as well as random initial object poses by adding independent uniform noise of up to 10 cm along each axis relative to the annotated positions and applying a uniformly sampled angular perturbation within $\pm$60 degrees around the annotated orientation. The task is considered successful if the laptop is opened to at least 95\% of its full opening angle within a specified time duration.

\textbf{Toilet} This task requires the robot to open a toilet lid. This task introduces randomness in the toilet assets, including variations in size and geometry and diverse lid shapes, as well as random initial object poses 
by adding independent uniform noise of up to 20 cm along each axis relative to the annotated positions and applying a uniformly sampled angular perturbation within $\pm$45 degrees around the annotated orientation. The task is considered successful if the toilet lid is opened to at least 95\% of its full opening angle within a specified time duration.

\subsection{Adroit Benchmarks}
We follow the original state space design used in prior RL benchmark work for these four representative dexterous manipulation tasks~\cite{rajeswaran2017learning} in MuJoCO simulator~\cite{Todorov2012MUJuCO}, where the Adroit robot hand’s proprioception space varies across tasks and is described separately below. Also, since the original benchmark uses joint torques as the action space, which is less commonly adopted in modern imitation learning approaches, we instead adopt another work that uses target joint positions as the action space, which achieve comparable task performance across all tasks~\cite{han2024learning}. We use the provided expert checkpoints to generate rollouts and collect successful trajectories, consisting of visual observations, observed robot joints, and target robot joints. We also adopt the original sampling strategies during data collection and the success criteria for imitation learning policy evaluation, which are detailed below.

\textbf{Door opening} This task requires the robot to undo the latch and drag the door open. The floating robot hand base can translate along the direction perpendicular to the door plane and rotate freely, and all 24 hand joints are fully actuated, resulting in a 28-dimensional robot proprioception space. This task introduces randomness in the initial door positions by uniformly sampling the Cartesian coordinates as $x \in \mathcal{X} \sim \mathcal{U}(-0.3,0)$, $y \in \mathcal{Y} \sim \mathcal{U}(0.2,0.35)$, and $z \in \mathcal{Z} \sim \mathcal{U}(0.252,0.402)$ (unit: m). The task is considered successful if the door opening angle is larger than 1.35 rad within a specified time duration.

\textbf{Tool use} This task requires the robot to pick up the hammer to drive the nail into the board. The floating robot hand base can rotate along the $x$ and $y$ axis, and all 24 hand joints are fully actuated, resulting in a 26-dimensional robot proprioception space. This task introduces randomness in the target nail heights by uniformly sampling the nail height as $h \in \mathcal{H} \sim \mathcal{U}(0.1,0.25)$ (unit: m). The task is considered successful if the Euclidean distance between the nail position and the goal nail position is smaller than 0.01 within a specified time duration.

\textbf{In-hand reorientation} This task requires the robot to reorient a pen to a goal orientation. The floating robot hand base is fixed, and only the 24 hand joints are actuated, resulting in a 24-dimensional robot proprioception space. This task introduces randomness in the goal pen orientation by uniformly sampling the pitch ($\alpha$) and yaw ($\beta$) angles as $\alpha \in \mathcal{A} \sim \mathcal{U}(-1,1)$ and $\beta \in \mathcal{B} \sim \mathcal{U}(-1,1)$ (unit: rad). At each time step $t$, if $\mathrm{o}^{\text{goal}} \cdot \mathrm{o}^{\text{pen}}_t > 0.90$ ($\mathrm{o}^{\text{goal}} \cdot \mathrm{o}^{\text{pen}}_t$ measures orientation similarity), then we have $\rho(t) = 1$. The task is considered successful if $\sum_{t=1}^T \rho(t) > 10$ ($T=100$).

\textbf{Object relocation} This task requires the robot to move a ball to a target position. The floating robot hand base can move freely, and all 24 hand joints are fully actuated, resulting in a 30-dimensional robot proprioception space. This task introduces randomness in the target positions by uniformly sampling the Cartesian coordinates as  $x \in \mathcal{X} \sim \mathcal{U}(-0.15,0.15)$, $y \in \mathcal{Y} \sim \mathcal{U}(-0.15,0.15)$, and $z \in \mathcal{Z} \sim \mathcal{U}(0.15,0.35)$ (unit: m). At each time step $t$, if $\sqrt{|\mathrm{p}^{\text{target}} - \mathrm{p}^{\text{ball}}_t|^2} < 0.10$, then we have $\rho(t) = 1$. The task is considered successful if $\sum_{t=1}^T \rho(t) > 10$ ($T=100$).

\subsection{Real-world Tasks}


We evaluate on four real-world dexterous manipulation tasks: \textit{Flower arrangement}, \textit{Chips pouring}, \textit{Dinner serving}, and \textit{Whiteboard erasing}. The robot proprioception space is shared across all four tasks and consists of the 7-DoF Franka Research 3 arm joint positions together with the 16-DoF LEAP-V1 left-hand finger joints (23 dimensions in total).

For all tasks, we collect demonstrations via a kinesthetic teach-and-replay pipeline and record synchronized RGB observations from a fixed third-person webcam. We further post-process the data to align the initial robot joint configurations across trajectories. The tasks are then performed under varying object configurations by relocating the target object within the workspace between rollouts. The success criteria are defined as:
\begin{itemize}
  \item \textit{Flower arrangement}: the robot grasps the bouquet from its initial location and inserts it into a vase.
  \item \textit{Chips pouring}: the robot grasps the chips can, lifts and tilts it over a target container to pour chips.
  \item \textit{Dinner serving}: the robot grasps the handle of a frying pan, lifts it, and transfers it to the plate location.
  \item \textit{Whiteboard erasing}: the robot grasps a cloth and wipes it across a marked whiteboard region, clearing the markings.
\end{itemize}

\section{Baseline policy design}
\label{appendix:policy_designs}
We show the detailed baseline policy design in this section. 

\noindent \textbf{Diffusion}: Following the design choices of \cite{diffusion_policy}, we implement a CNN-based diffusion policy that represents the visuomotor mapping as a conditional denoising diffusion process. A 1D temporal Convolutional Neural Network with a U-Net-like structure serves as the noise-prediction network, which iteratively refines a sequence of actions from Gaussian noise during inference. To incorporate visual context, the model integrates pre-extracted flow or DynaMo features as global conditioning via Feature-wise Linear Modulation (FiLM) layers. Finally, the policy employs a receding horizon control strategy to ensure temporal action consistency and provide robustness against execution latency.

We implement the diffusion policy using the AdamW optimizer with a learning rate of $10^{-4}$, weight decay of $10^{-6}$, and betas of $[0.95, 0.999]$. The model is trained for 1000 epochs with a batch size of 256, incorporating a linear learning rate warmup of 500 steps followed by a cosine decay. The architecture utilizes an observation horizon ($T_o$) of 2. For the action prediction horizon ($T_p$) and the action execution horizon ($T_a$), we evaluate two configurations—$(T_a, T_p) \in \{(4, 8), (8, 16)\}$—to determine the optimal settings for each task (see Table~\ref{tab:Diffusion_arch}). For the diffusion process, we employ a DDPMScheduler with 100 training timesteps and a squared cosine noise schedule. The noise-prediction network features a 1D temporal U-Net with a kernel size of 5 and down-sampling dimensions of 256, 512, and 1024.

\noindent \textbf{ACT}: We follow the policy architecture design as~\cite{zhao2023learning}. During training, we first sample tuples of flow (or DynaMo) features and joint angles, together with the corresponding action sequence as prediction target. We then infer style variable $z$, modeled as  a diagonal Gaussian, using CVAE encoder. This variable is then fed into a decoder alongside visual features and current joint states. The decoder, structured as a Transformer with cross-attention, learns to reconstruct the action sequence from these inputs. At test time, the CVAE encoder is discarded, and the CVAE decoder is used as the policy. The latent variable $z$ is fixed to zero—the mean of the prior distribution. This allows the decoder to act as a deterministic policy that maps multi-modal observations directly to robot actions.

We train the ACT model with a learning rate of $10^{-5}$ and a batch size of 8. The architecture comprises a 4-layer Transformer encoder and a 7-layer decoder, featuring 8 attention heads, a hidden dimension of 512, and a feedforward dimension of 3200. We apply a dropout rate of 0.1 and set the KL divergence weight to 10. To optimize performance, we evaluated three chunk sizes—10, 20, and 40—and selected the optimal value for each task based on the results in Table~\ref{tab:ACT_archi}.

\noindent \textbf{UVA}: We follow the architecture of Unified Video Action (UVA)~\cite{li2025uva}, a Masked Autoregressive (MAR) Vision Transformer that jointly models future video frames and robot actions. A frozen KL-16 VAE first encodes each $256\times256$ image into a $16\times16\times16$ latent grid (256 tokens per frame). A 12-layer MAR encoder ingests the latent tokens of the conditioning frames together with masked future tokens, proprioceptive history, and the action sequence; a 12-layer MAR decoder then produces a shared latent representation. Two lightweight diffusion heads operate on this representation: a video head that denoises the masked future visual tokens to predict future frames, and an action head that denoises a spatially-compressed feature into an action chunk. At inference, we run UVA in its \textit{full dynamics mode}, in which both heads are active, so the model jointly predicts the future video and the action chunk at every query.

We use the official implementation with its default hyperparameters. The model is initialized from the pretrained \texttt{mar\_base} backbone ($\sim$262M trainable parameters) and trained for 300 epochs using the AdamW optimizer (learning rate $10^{-4}$, weight decay $0.02$, betas $[0.9, 0.95]$), a cosine learning-rate schedule with 1000 warmup steps, mixed-precision (fp16) training, and an exponential moving average of the weights. We condition on 4 frames and predict an action chunk of length 8; both diffusion heads have depth 6 and width 1024 and use 100 denoising steps. We do not perform additional architecture search for UVA due to the extensive training time required.

\noindent \textbf{KODex}: In the literature, Koopman-based methods commonly employ polynomial and sinusoidal (cosine/sine) lifting functions~\cite{han2023utility, shi2021acd, shi2026koopman}. As a result, the mapping between the original state and the lifted state is predefined, allowing the one-step prediction objective to be solved directly via least squares. However, this formulation typically does not support optimization with multi-step prediction losses. To optimize the predefined lifting functions, we design three variants of polynomial bases and report the corresponding dimensionalities in Table.~\ref{tab:lifting_dims}. Since the flow features and DynaMo features share the same input dimensionality, the resulting lifted dimensionalities are identical for both. The optimized Koopman matrix is a square matrix whose dimensionality matches the lifted state dimension for each task.
\begin{table*}[t]
    \centering
    \caption{Lifting dimensionalities for three lifting variants across seven simulation tasks}
    \label{tab:lifting_dims}
        \resizebox{0.98\textwidth}{!}{%
    \begin{tabular}{c|c|c|c|c}
        \hline
        \multirow{3}{*}{\diagbox[width=9em,height=2.8\line]{Variant}{Task Group}} 
        & \multirow{3}{*}{Bucket / Laptop / Toilet / Door} 
        & \multirow{3}{*}{Tool} 
        & \multirow{3}{*}{Reorientation} 
        & \multirow{3}{*}{Relocation} \\
        & & & & \\
        & $n_h{=}28, n_o{=}128$ & $n_h=26, n_o{=}256$ & $n_h=24, n_o{=}256$ & $n_h=30, n_o{=}256$ \\
        \hline
        V1 
        & 846 & 1171 & 1116 & 1293 \\
        \hline
        V2
        & 718 & 915 & 860 & 1037 \\
        \hline
        V3
        & 1030 & 1479 & 1420 & 1609 \\
        \hline
    \end{tabular}
    }
    \vspace{0.5em}
    \footnotesize{
      The Tool, Reorientation, and Relocation tasks use goal-conditioned policies, so their visual state dimensionality is 256.
    }
\end{table*}

Let $\mathbf{x}_h \in \mathbb{R}^{n_h}$ denote the original robot state
and $\mathbf{x}_o \in \mathbb{R}^{n_o}$ denote the original visual states. All lifting variants include the original states and their element-wise
second-order terms for both the hand and visual states, i.e.,
\[
\Phi_{\text{base}}(\mathbf{x}_h, \mathbf{x}_o)
=
\big[
\mathbf{x}_h,\;
\mathbf{x}_h^{\odot 2},\;
\mathbf{x}_o,\;
\mathbf{x}_o^{\odot 2}
\big],
\]
where $\odot$ denotes element-wise exponentiation.
\newline
\textbf{V1}: We further augments the base lifting with third-order
polynomial terms of the visual states, pairwise interaction terms between
robot states, and third-order polynomial terms of the robot states:
\[
\Phi_{\text{v1}} =
\Phi_{\text{base}}
\;\cup\;
\big[
\mathbf{x}_o^{\odot 3},\;
\{x_{h,i} x_{h,j}\}_{i<j},\;
\mathbf{x}_h^{\odot 3}
\big].
\]
\noindent \textbf{V2}: We remove higher-order polynomial terms of the visual states
and retains only nonlinear expansions of the robot states:
\[
\Phi_{\text{v2}} =
\Phi_{\text{base}}
\;\cup\;
\big[
\{x_{h,i} x_{h,j}\}_{i<j},\;
\mathbf{x}_h^{\odot 3}
\big].
\]
\noindent \textbf{V3}: We extend V2 by additionally incorporating
trigonometric features for both robot and visual states:
\[
\Phi_{\text{v3}} =
\Phi_{\text{v2}}
\;\cup\;
\big[
\cos(\mathbf{x}_h),\;
\sin(\mathbf{x}_h),\;
\cos(\mathbf{x}_o),\;
\sin(\mathbf{x}_o)
\big].
\]
\newline
\textbf{KOROL}: In the vanilla implementation of KOROL, a pretrained ResNet-18 visual encoder is fine-tuned every epoch, while the Koopman operator is updated at fixed intervals. Robot states and 8-dimensional visual object features extracted from images are first lifted using hand-crafted polynomial basis functions, and these lifted representations are used to alternately update the feature extractor and re-estimate the Koopman matrix via least-squares optimization. However, for our implementation of KOROL, we replace the ResNet-18 model with Flow and DynaMo feature extractors and keeps the visual backbone frozen. Instead, we introduce a learnable MLP projection head to process the visual features, which is updated every epoch to adapt features for Koopman learning. 

We retain an output feature dimension of 8, use a learning rate of 1e-4 for Flow features, and tune the learning rate to 0.025 for DynaMo features for stable convergence. The prediction horizon is fixed to 15, consistent with other baselines. We design three MLP variants with varying hidden dimensions for model optimization as shown in 
Table \ref{tab:korol_archi}.
\section{Baseline design optimization}
\label{appendix:layer_size}
For the baseline models, we make substantial efforts to optimize their architectures to ensure a fair comparison. Specifically, for diffusion-based models, we tune the combination of execution horizon and prediction horizon, which is critical to their performance. For ACT models, we tune the prediction horizon. For KODex models, we select appropriate polynomial lifting functions, and for KOROL models, we tune the MLP used to ingest visual features. We generated five random seeds for parameter initialization per architecture choice, per baseline, and per task, except for the KODex method. For each baseline policy, we report the mean and standard deviation of the task success rate on the test set over five random seeds. We then select the best-performing configuration based on average performance for comparison with our method in Section~\ref{sec:general_eval}. For diffusion-based models, Tables~\ref{tab:Diffusion_arch} report results using flow features and DynaMo features, respectively. Similarly, for ACT models, refer to Tables~\ref{tab:ACT_archi}; for KODex models, refer to Tables~\ref{tab:KODex_archi}; and for KOROL models, refer to Tables~\ref{tab:korol_archi}.

\begin{table*}[t]
    \centering
    \caption{Policy Architecture Optimization for Diffusion Models}
    \label{tab:Diffusion_arch}
    \resizebox{0.98\textwidth}{!}{%
    \begin{tabular}{c|cc|cc|cc|cc|cc|cc|cc|cc}
        \hline
        \multirow{2}{*}{Architecture}
        & \multicolumn{16}{c}{Success Rate (\%, $\uparrow$)} \\
        \cline{2-17}
        & \multicolumn{2}{c|}{Bucket}
        & \multicolumn{2}{c|}{Laptop}
        & \multicolumn{2}{c|}{Toilet}
        & \multicolumn{2}{c|}{Door}
        & \multicolumn{2}{c|}{Hammer}
        & \multicolumn{2}{c|}{Pen}
        & \multicolumn{2}{c|}{Relocation}
        & \multicolumn{2}{c}{Average}\\
        $(T_o, T_a, T_p)$ & Flow & DynaMo & Flow & DynaMo & Flow & DynaMo & Flow & DynaMo & Flow & DynaMo & Flow & DynaMo & Flow & DynaMo & Flow & DynaMo \\
        \hline
        (2, 4, 8)
        & 90.7 $\pm$ 3.9 & 75.3 $\pm$ 5.0
        & 78.0 $\pm$ 4.5 & 78.7 $\pm$ 6.5
        & 69.0 $\pm$ 3.1 & 68.0 $\pm$ 8.8
        & 0.0  $\pm$ 0.0 & 0.0  $\pm$ 0.0
        & 88.3 $\pm$ 2.9 & 65.3 $\pm$ 8.6
        & 80.7 $\pm$ 2.5 & 32.7 $\pm$ 3.8
        & 89.2 $\pm$ 6.0 & 50.0 $\pm$ 2.1
        & \textbf{70.8 $\pm$ 30.5} & \textbf{52.7 $\pm$ 25.5} \\
        \hline
        (2, 8, 16)
        & 92.7 $\pm$ 2.5 & 76.0 $\pm$ 3.3
        & 60.0 $\pm$ 19.1 & 82.7 $\pm$ 4.9
        & 67.6 $\pm$ 4.6 & 76.7 $\pm$ 4.7
        & 0.0  $\pm$ 0.0 & 0.0  $\pm$ 0.0
        & 92.7 $\pm$ 1.3 & 78.7 $\pm$ 6.2
        & 81.3 $\pm$ 4.5 & 3.3  $\pm$ 2.9
        & 86.7 $\pm$ 2.9 & 55.3 $\pm$ 3.4
        & 68.9 $\pm$ 30.6 & 53.2 $\pm$ 25.2 \\
        \hline
    \end{tabular}
    }
\end{table*}

\begin{table*}[t]
    \centering
    \caption{Policy Architecture Optimization for ACT Models}
    \label{tab:ACT_archi}
    \resizebox{0.98\textwidth}{!}{%
    \begin{tabular}{c|cc|cc|cc|cc|cc|cc|cc|cc}
        \hline
        \multirow{2}{*}{Model Architecture}
        & \multicolumn{16}{c}{Success Rate (\%, $\uparrow$)} \\
        \cline{2-17}
        & \multicolumn{2}{c|}{Bucket}
        & \multicolumn{2}{c|}{Laptop}
        & \multicolumn{2}{c|}{Toilet}
        & \multicolumn{2}{c|}{Door}
        & \multicolumn{2}{c|}{Tool}
        & \multicolumn{2}{c|}{Reorientation}
        & \multicolumn{2}{c|}{Relocation}
        & \multicolumn{2}{c}{Average}\\
        (chunk size) & Flow & DynaMo & Flow & DynaMo & Flow & DynaMo & Flow & DynaMo & Flow & DynaMo & Flow & DynaMo & Flow & DynaMo & Flow & DynaMo \\
        \hline
10
& 80.7 $\pm$ 5.7 & 72.7 $\pm$ 8.5
& 90.0 $\pm$ 3.7 & 78.7 $\pm$ 2.7
& 74.5 $\pm$ 6.4 & 82.7 $\pm$ 3.9
& 0.0  $\pm$ 0.0 & 0.0 $\pm$ 0.0
& 52.0 $\pm$ 18.3 & 88.7 $\pm$ 8.4
& 80.0 $\pm$ 7.7  & 28.0 $\pm$ 8.7
& 82.0 $\pm$ 9.8 & 72.0 $\pm$ 3.8
& 65.6 $\pm$ 30.3  & 60.4 $\pm$ 27.0 \\
\hline
20
& 77.3 $\pm$ 5.5 & 83.3 $\pm$ 7.0
& 86.7 $\pm$ 4.1 & 92.0 $\pm$ 3.4
& 69.0 $\pm$ 8.1 & 78.7 $\pm$ 3.4
& 23.2 $\pm$ 9.3 & 41.3 $\pm$ 5.1
& 79.2 $\pm$ 11.6 & 96.7 $\pm$ 4.1
& 85.0 $\pm$ 5.8  & 25.3 $\pm$ 4.5
& 89.4 $\pm$ 7.8  & 72.0 $\pm$ 3.0
& \textbf{73.4 $\pm$ 25.7}  & \textbf{69.9 $\pm$ 24.7} \\
\hline
40
& 74.7 $\pm$ 6.9 & 74.7 $\pm$ 2.7
& 83.3 $\pm$ 5.8 & 76.0 $\pm$ 6.5
& 68.3 $\pm$ 4.5 & 82.0 $\pm$ 5.4
& 44.2 $\pm$ 8.0 & 44.7 $\pm$ 13.9
& 67.2 $\pm$ 13.9 & 92.0 $\pm$ 1.8
& 78.8 $\pm$ 5.2  & 26.0 $\pm$ 8.0
& 84.6 $\pm$ 3.9  & 81.3 $\pm$ 1.8
& 71.6 $\pm$ 18.7  & 68.1 $\pm$ 22.7 \\

        \hline
    \end{tabular}
    }
\end{table*}

\begin{table*}[t]
    \centering
    \caption{Policy Architecture Optimization for KODex Models}
    \label{tab:KODex_archi}
    \resizebox{0.98\textwidth}{!}{%
    \begin{tabular}{c|cc|cc|cc|cc|cc|cc|cc|cc}
        \hline
        \multirow{2}{*}{Model Architecture}
        & \multicolumn{16}{c}{Success Rate (\%, $\uparrow$)} \\
        \cline{2-17}
        & \multicolumn{2}{c|}{Bucket}
        & \multicolumn{2}{c|}{Laptop}
        & \multicolumn{2}{c|}{Toilet}
        & \multicolumn{2}{c|}{Door}
        & \multicolumn{2}{c|}{Tool}
        & \multicolumn{2}{c|}{Reorientation}
        & \multicolumn{2}{c|}{Relocation}
        & \multicolumn{2}{c}{Average}\\
        & Flow & DynaMo & Flow & DynaMo & Flow & DynaMo & Flow & DynaMo & Flow & DynaMo & Flow & DynaMo & Flow & DynaMo & Flow & DynaMo \\
        \hline
        V1
        & 10.0 & 80.0
        & 6.7 & 33.3
        & 3.4 & 16.7
        & 3.3 & 50.0
        & 36.7 & 20.0 & 73.3 & 70.0 & 26.7 & 23.3  & 22.9 $\pm$23.7 & 41.9 $\pm$23.4 \\
        \hline
        V2
        & 13.3 & 80.0
        & 10.0 & 33.3
        & 10.3 & 6.7
        & 10.0 & 53.3
        & 30.0 & 33.3 & 76.7 & 73.3 & 23.3 & 46.7 & 24.8 $\pm$22.4 & \textbf{46.7 $\pm$23.3} \\
        \hline
        V3
        & 20.0 & 86.7
        & 23.3 & 13.3
        & 3.4 & 3.3
        & 13.3 & 53.3
        & 30.0 & 50.0 & 63.3 & 76.7 & 33.3 & 20.0 & \textbf{26.7 $\pm$17.6} & 43.3 $\pm$29.7 \\
        \hline
    \end{tabular}
    }
\end{table*}

\begin{table*}[t]
    \centering
    \caption{Policy Architecture Optimization for KOROL Models}
    \label{tab:korol_archi}
    \resizebox{0.98\textwidth}{!}{%
    \begin{tabular}{c|cc|cc|cc|cc|cc|cc|cc|cc}
        \hline
        \multirow{2}{*}{Model Architecture}
        & \multicolumn{16}{c}{Success Rate (\%, $\uparrow$)} \\
        \cline{2-17}
        & \multicolumn{2}{c|}{Bucket}
        & \multicolumn{2}{c|}{Laptop}
        & \multicolumn{2}{c|}{Toilet}
        & \multicolumn{2}{c|}{Door}
        & \multicolumn{2}{c|}{Tool}
        & \multicolumn{2}{c|}{Reorientation}
        & \multicolumn{2}{c|}{Relocation}
        & \multicolumn{2}{c}{Average}\\
         (hidden dim) & Flow & DynaMo & Flow & DynaMo & Flow & DynaMo & Flow & DynaMo & Flow & DynaMo & Flow & DynaMo & Flow & DynaMo & Flow & DynaMo \\
        \hline
        $[64]$
        & 50.7 $\pm$16.2 & 56.7 $\pm$27.6
        & 20.7 $\pm$3.9 & 72.0 $\pm$3.4
        & 33.8 $\pm$8.6 & 42.8 $\pm$10.6
        & 10.7 $\pm$6.8  & 11.33 $\pm$2.67
        & 32.0 $\pm$17.2 & 13.3 $\pm$12.11 & 32.0 $\pm$20.1 & 52.0 $\pm$9.1 & 8.0$\pm$ 9.1 & 15.33 $\pm$4.52  & \textbf{26.8 $\pm$13.8} & 37.6 $\pm$22.5 \\
        \hline
        $[2\cdot\text{input\_dim},\,128]$
        & 46.7 $\pm$7.6 & 81.3 $\pm$4.5
        & 19.3 $\pm$11.0 & 71.3 $\pm$1.6
        & 27.6 $\pm$3.1 & 43.4 $\pm$6.8
        & 9.3 $\pm$3.9 &  9.33 $\pm$2.49
        & 32.7 $\pm$17.2 & 4.7 $\pm$6.2 & 36.7 $\pm$11.2 & 46.0 $\pm$6.11 & 8.0 $\pm$2.7 & 15.33 $\pm$3.4  & 25.8 $\pm$ 13.3 & 44.9 $\pm$ 24.4\\
        \hline
        $[512, 256, 128]$
        & 35.3 $\pm$20.9 & 78.7 $\pm$4.5
        & 16.9 $\pm$5.7 & 71.3 $\pm$6.2
        & 16.6 $\pm$13.7 & 53.1 $\pm$8.3
        & 14.0 $\pm$4.4 &  11.33 $\pm$1.63
        & 13.0 $\pm$2.7 & 5.33 $\pm$9.09 & 34.0 $\pm$17.6 & 37.33 $\pm$5.33 & 6.7 $\pm$2.1 & 17.33 $\pm$3.89  & 19.5 $\pm$10.1 & \textbf{46.1 $\pm$23.6} \\
        \hline
    \end{tabular}
    }
\end{table*}
\section{Hyper-parameter for Koopman Learning}
\label{appendix:learning_detail}
As shown in Table.~\ref{tab:hyper_param_our_method}, we use the same training hyperparameters for our method across all seven simulation tasks and four real-world tasks, with the only modification being an increased prediction horizon of 50 steps for real-world tasks, which typically involve trajectories longer than 150 steps. 
\begin{table*}[t]
    \centering
    \caption{Hyper-parameters on seven simulator tasks}
    \label{tab:hyper_param_our_method}
    \resizebox{0.98\textwidth}{!}{%
        \begin{tabular}{c|c|c|c|c|c|c}
            \hline
            Hidden Layer & Lifting Dimension & Activation & Encoder lr & Koopman lr & Multi-step Length & Identity Koopman Matrix Initialization \\
            \hline
            (128, 256) & 256 & ReLU & 5e-4 & 5e-5 & 15 & True \\
            \hline
        \end{tabular}
    }
\end{table*}

\section{Dataset Modification}
\label{sec:data_appendix}
In typical imitation learning problems, the first action
\(\mathrm{a}_1^{(i)}\) varies across demonstrations and is therefore unavailable at inference time.
To address this issue, UBMs utilize an auxiliary initial action
\(\mathrm{a}_0^{(i)} = \mathrm{q}_1^{(i)}\),
where \(\mathrm{q}_1^{(i)}\) denotes the initial robot joint configuration. In practice, it is reasonable to initialize the robot to the same configuration across demonstrations, so
\(\mathrm{q}_1^{(i)}\) (and thus \(\mathrm{a}_0^{(i)}\)) is identical for all \(i\). To ensure consistent trajectory lengths during learning, we also duplicate the initial image by setting
\(\mathrm{I}_0^{(i)} = \mathrm{I}_1^{(i)}\). This process results in a new dataset
$
\tilde{\mathcal{D}}
=
\Big[
\{\mathrm{I}^{(1)}_t, \mathrm{a}^{(1)}_t\}_{t=0}^{T^{(1)}},
\;\ldots,\;
\{\mathrm{I}^{(N)}_t, \mathrm{a}^{(N)}_t\}_{t=0}^{T^{(N)}}
\Big].
$
Our goal is to learn a UBM from the dataset $\tilde{\mathcal{D}}$ that will enable the robot to autonomously perform the associated dexterous skill. 
\section{Evaluation of Visual Feature Prediction}
\label{appendix:visual_fea_pred}
\begin{figure}[t]
\centering
\includegraphics[width=0.9\textwidth]{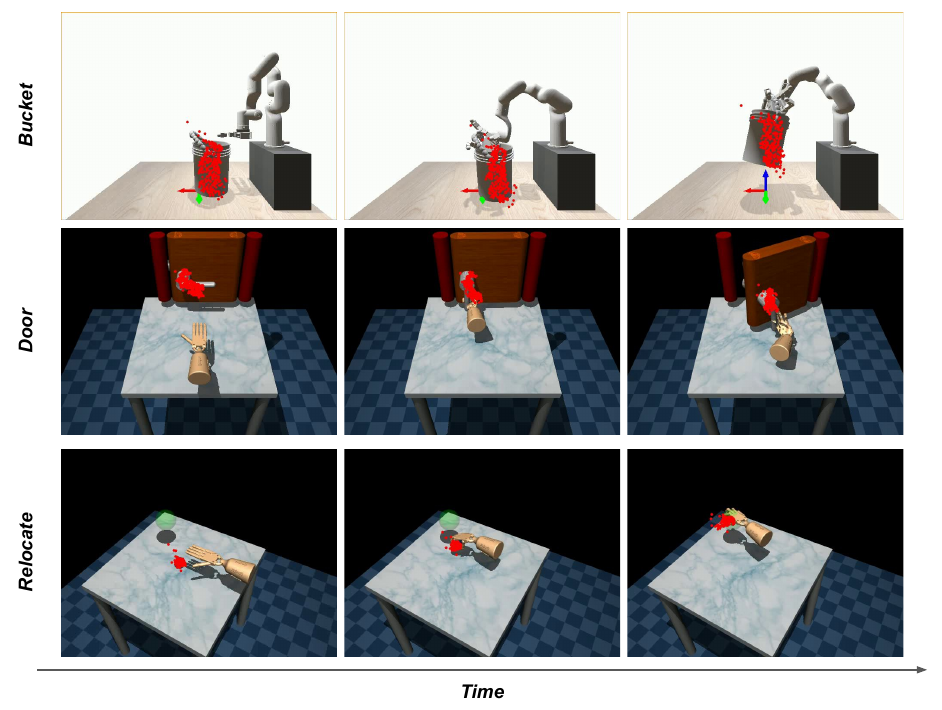}
\caption{We visualize the predicted flow points by decoding the flow features and overlaying them on the corresponding images.}
\label{fig:flow_qualitative}
\end{figure}

A key benefit of our method is the joint prediction of both robot states and manipulation-centric environment dynamics. The accuracy of robot state prediction has already been demonstrated in the previous section through task performance. In this section, we evaluate the prediction accuracy of manipulation-centric environment dynamics, which can be utilized in multiple ways, such as triggering replanning (see Appendix.~\ref{appendix:Reactivity}), serving as an inference-time verification signal (see result discussion of this section), or acting as an object-centric tracking reward for reinforcement learning~\cite{han2024learning}. Since Diffusion, ACT, and KOROL do not support explicit object prediction, and KODex performs significantly worse as shown in Section~\ref{sec:general_eval}, we do not include comparisons with these baseline methods.
\newline
\textit{Flow Feature Prediction} First, in Fig.~\ref{fig:flow_qualitative}, we present qualitative results for three reconstructed flow point trajectories on the test set for the Bucket, Door, and Relocate tasks. These trajectories are obtained by decoding the flow features predicted via Koopman rollouts using the flow decoder learned during autoencoder training. We further visualize the reconstructed flow points by overlaying them on the corresponding images as red dots. From these results, we can see that the predicted flow features effectively capture and reflect the manipulation objectives throughout the entire task execution.

\begin{figure*}[t]
\centering
\includegraphics[width=0.95\textwidth]{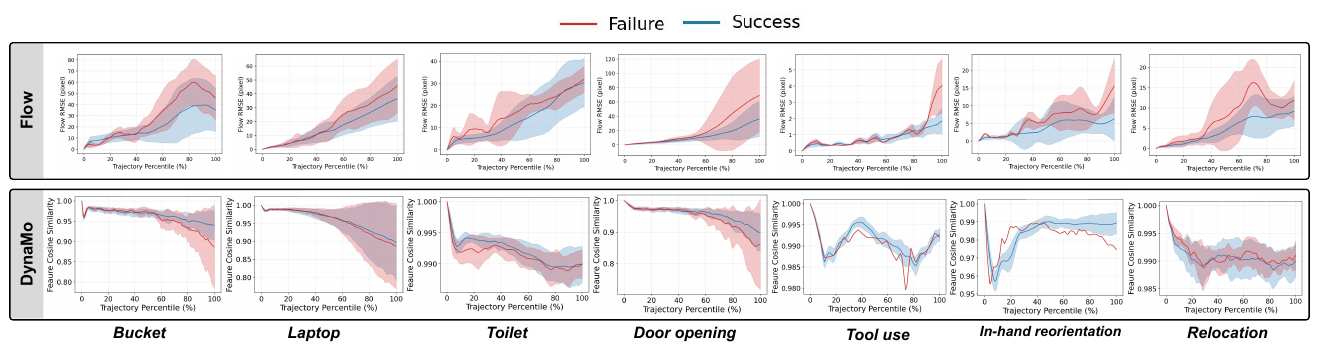}
\caption{In the top row, we show the reconstruction error (RMSE) between the predicted and ground-truth flow points. In the bottom row, we report the cosine similarity between the predicted DynaMo features and the ground-truth trajectories.}
\label{fig:flow_prediction}
\end{figure*}

Furthermore, we evaluate the quantitative prediction quality of the reconstructed flow points. In Fig.~\ref{fig:flow_prediction} (top row), we report the root mean square error (RMSE) between the predicted and ground-truth flow points over time for all seven simulation tasks on the test sets, averaged across all test rollouts. Since rollout lengths vary across runs, we visualize the error using normalized trajectory percentiles. We further separate the prediction errors for successful and failed rollouts. As shown in the figure, failed rollouts consistently exhibit larger flow prediction errors than successful ones, revealing a strong correlation between degraded object flow prediction and poor robot execution. 

\begin{figure}[t]
\centering
\includegraphics[width=0.48\textwidth]{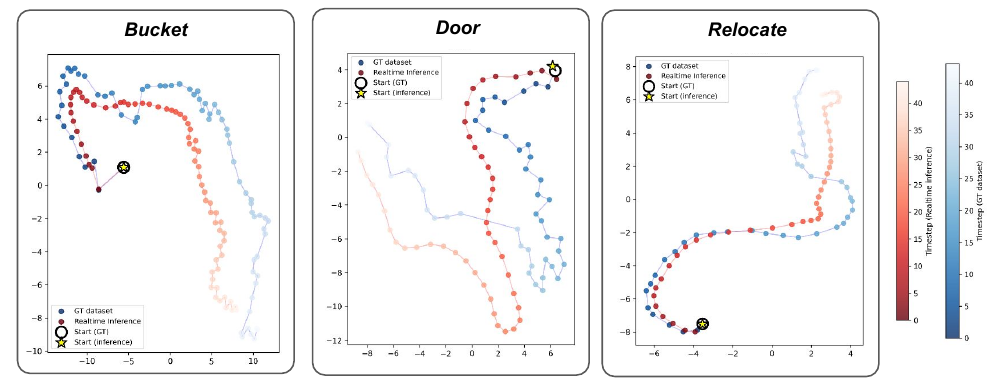}
\caption{We visualize the predicted DynaMo features by projecting them together with the ground-truth features using t-SNE.}
\label{fig:dynamo_qualitative}
\end{figure}

\noindent \textit{Dynamo Feature Prediction} For DynaMo features, following an evaluation procedure similar to that used for flow features, we first present three qualitative results comparing the predicted and ground-truth DynaMo features in the latent space using t-SNE~\cite{maaten2008visualizing} visualization, as shown in Fig.~\ref{fig:dynamo_qualitative}. From these results, we observe that the predicted high-dimensional visual features align well with the ground-truth features.

We then evaluate the cosine similarity between the predicted and ground-truth features for all seven simulation tasks on the test sets, averaging the results across all test rollouts and visualizing them using trajectory percentiles, as shown in the bottom row of Fig.~\ref{fig:flow_prediction}. Overall, we observe that the cosine similarity generally decreases over time, indicating increasing prediction error, which is expected for long-horizon predictions. Notably, for the \textit{Tool Use} and \textit{In-Hand Reorientation} tasks, the feature similarity varies only slightly, resulting in non-monotonic curves. 

In addition, although the trend is less pronounced than in the flow prediction curves, we still observe a similar pattern: failed rollouts generally exhibit lower cosine similarity than successful ones.
\section{Evaluation of Online Reactivity}
\label{appendix:Reactivity}
In this section, we illustrate how K-UBMs can achieve \textbf{system-level online reactivity}, while remaining capable of open-loop rollouts under nominal conditions. We evaluate K-UBM’s ability to adapt and react to unexpected environmental changes during execution across all seven simulation tasks, comparing it against diffusion- and ACT-based policies, all using flow-based observations.  

First, we elaborate on the replanning experiments for K-UBM. During execution, K-UBM compares the observed object behavior with the predicted nominal object behavior and triggers replanning only when the deviation is sufficiently large. Therefore, K-UBM is well suited for handling \textit{considerable deviations}, which can be detected more reliably. For each task, we introduce a nontrivial perturbation once during the rollout (see Table.~\ref{table:perturbation_range} for the per-task perturbation ranges). K-UBM is expected to autonomously detect the perturbation and generate a new trajectory for switching, i.e., first interpolating from the current state to the closest point on the generated trajectory and then tracking the new trajectory to complete the task after the perturbation. 

\begin{table}[h]
\centering
\resizebox{0.8\textwidth}{!}{
\begin{tabular}{lcc}
\toprule
\textbf{Task} & \textbf{Perturbation Type} & \textbf{Perturbation Range} \\
\midrule
Bucket & Object position $\Delta p$ & $0.065 \pm 0.029$ m \\
Laptop & Object position $\Delta p$ & $0.099 \pm 0.042$ m \\
Toilet & Object position $\Delta p$ & $0.191 \pm 0.071$ m \\
Door & Object position $\Delta p$ & $0.181 \pm 0.046$ m \\
Tool & Goal position $\Delta p_g$ & $0.062 \pm 0.037$ m \\
In-hand Reorientation & Goal orientation $\Delta R_g$ & $58.3 \pm 23.8^\circ$ \\
Relocation & Goal position $\Delta p_g$ & $0.361 \pm 0.143$ m \\
\bottomrule
\end{tabular}
}
\caption{Perturbation ranges used in the reactivity experiments. $\Delta p$ and $\Delta p_g$ denote object and goal position perturbations, respectively, measured in meters. $\Delta R_g$ denotes goal orientation perturbation, measured in degrees.}
\label{table:perturbation_range}
\end{table}

During experiments, we continuously track the visual features and compute the discrepancy between the predicted ones and those observed from real-time images. When this error increases sharply (i.e., changes in the flow centroid position for the flow-based policy), a replanning trigger is activated, prompting the policy to generate a new trajectory. See the left block of Fig.~\ref{fig:flow_replan_trigger} in Appendix.~\ref{sec:Replan_trigger_detection} for two examples, illustrating that when the environment changes, the resulting increase in prediction error is pronounced and readily detectable. For the diffusion- and ACT-based policies, since they are reactive by nature, we simply continue querying the policy after the perturbation using the updated flow observations.

\begin{table}[h]
\centering
\resizebox{1.0\textwidth}{!}{%
\begin{tabular}{l ccccccc|c}
\toprule
& \textbf{Bucket} 
& \textbf{Laptop} 
& \textbf{Toilet} 
& \textbf{Door} 
& \textbf{Tool} 
& \textbf{In-hand Reori} 
& \textbf{Reloc} 
& \textbf{Average} \\
\midrule

Diffusion 
& SR: 23 / 30 
& SR: 22 / 30
& SR: 13 / 30
& SR: 0 / 30
& SR: 12 / 30
& SR: 1 / 30 
& SR: 16 / 30 
& SR: 0.41($\pm$0.28)
\\

ACT 
& SR: 24 / 30
& SR: 23 / 30
& SR: 11 / 30
& SR: 10 / 30 
& SR: 9 / 30 
& SR: 0 / 30
& SR: 10 / 30 
& SR: 0.41($\pm$0.26)
\\

K-UBM
& \begin{tabular}{c}
RT: 22 / 30 \\
SR: 15 / 30
\end{tabular}
& \begin{tabular}{c}
RT: 27 / 30 \\
SR: 17 / 30
\end{tabular}
& \begin{tabular}{c}
RT: 17 / 30 \\
SR: 07 / 30
\end{tabular} & \begin{tabular}{c}
RT: 30 / 30 \\
SR: 15 / 30
\end{tabular} & \begin{tabular}{c}
RT: 29 / 30 \\
SR: 23 / 30
\end{tabular} & \begin{tabular}{c}
RT: 30 / 30 \\
SR: 17 / 30
\end{tabular} & \begin{tabular}{c}
RT: 30 / 30 \\
SR: 27 / 30
\end{tabular}
& \begin{tabular}{c}
RT: 0.88($\pm$0.16) \\
SR: \textbf{0.58($\pm$0.20)}
\end{tabular} \\
\bottomrule
\end{tabular}
}
\caption{
\textbf{RT} and \textbf{SR}: trigger and task success ratios. \textbf{RT} is used to evaluate whether K-UBM can successfully detect environmental changes by comparing the real-time perception with the predicted nominal trajectory.}
\label{tab:reactivity_success_count}
\end{table}

K-UBM demonstrates competitive replanning performance, achieving the highest average success rate across all 210 instances, with 30 rollouts per task (see Table.~\ref{tab:reactivity_success_count}). This is achieved by K-UBM’s ability to replan from the beginning, which provides strong reactivity under environmental perturbations. In contrast, diffusion- and ACT-based policies need to learn such reactive behavior purely from data, which can make essential recovery behaviors after environmental changes more likely to be out of distribution, leading to worse task success rates.

We also analyze the failures of K-UBM's replanning and identify two modes: i) Trigger Failure: the 2D flow-based heuristic for detecting environmental changes can be less sensitive to certain 3D deviations, leading to missed triggers when using a task-invariant threshold. ii) Replanning Failure: effective replanning may require post-trigger transition behaviors unobserved in the data (e.g., recovering after toppling an object, or collision avoidance).

\begin{figure}[t]
\centering
\includegraphics[width=0.8\textwidth]{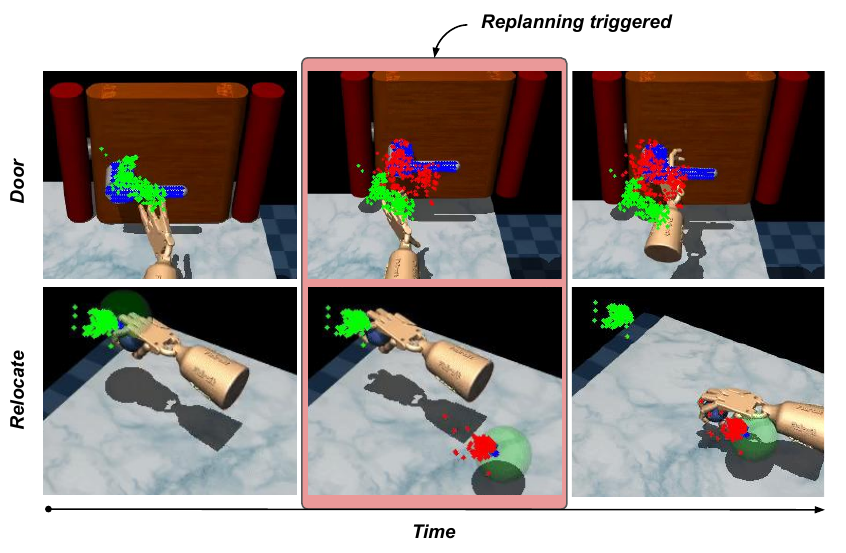}
\caption{For both tasks, the left column shows the robot executing the original trajectory, where green points denote the predicted flow points and blue points denote the ground-truth flow points. The middle column illustrates the environment change. Specifically, a new handle position for the door-opening task and a new goal position for the relocation task—which leads to a large discrepancy between the predicted (green) and ground-truth (blue) flow points. The right column shows the robot following the replanned trajectory, with red points indicating the new flow points.}
\label{fig:replan_trigger_flow_qualitative}
\end{figure}

Lastly, we present two qualitative examples to illustrate the replanning procedure in Fig.~\ref{fig:replan_trigger_flow_qualitative}. These results show that once replanning is triggered, the framework can reliably detect the change and seamlessly switch the robot to a replanned trajectory to successfully complete the updated task.

\section{Key design choices for learning Koopman based UBMs}
\label{sec:Recipes}
In this section, we systematically examine the key design choices that enable effective Koopman model learning. Through extensive experiments, we identify two critical factors: i) identity matrix initialization of the Koopman matrix, and ii) gradient clipping and the use of smaller learning rates for the Koopman matrix.

Intuitively, temporally neighboring states should be similar. Based on this, we initialize the Koopman matrix as the identity matrix. Through experiments, we find that this choice is crucial, as multi-step predictions from a randomly initialized Koopman matrix can produce large prediction errors, leading to unstable gradient descent.

Additionally, we find that enabling gradient clipping and using separate learning rates for the encoder and the Koopman matrix are beneficial, as multi-step matrix rollouts accumulate gradient signals during prediction, which can otherwise lead to unstable updates.

\begin{figure*}[ht]
\centering
\includegraphics[width=0.95\textwidth]{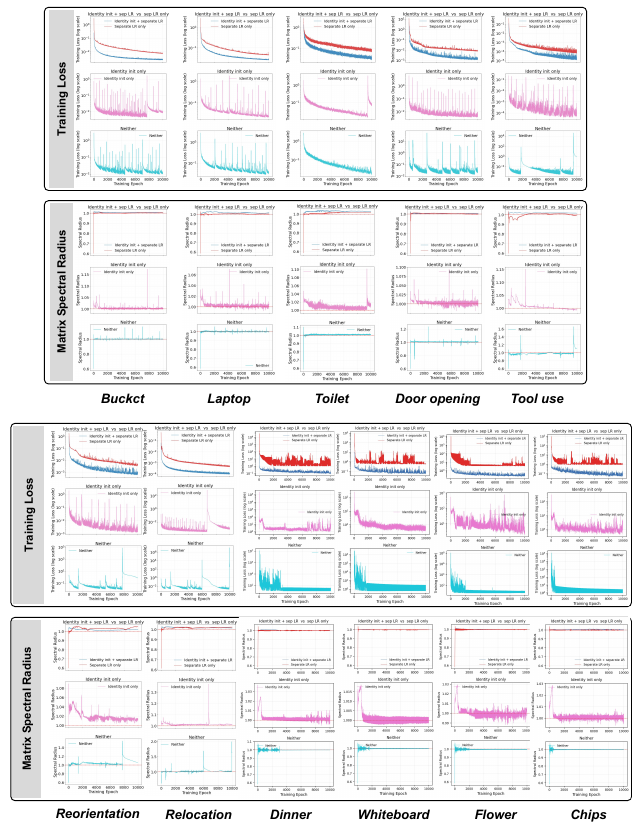}
\caption{We conduct ablation studies to evaluate two key factors that enable efficient Koopman learning. \textit{Identity init + separate LR} denotes the training configuration used throughout the main paper, which incorporates both key factors. \textit{Separate LR only} corresponds to using separate learning rates for the lifting network and Koopman matrix without identity initialization. \textit{Identity init only} initializes the Koopman matrix as the identity but uses a shared learning rate, while \textit{Neither} disables both. From the results, it can be observed that without separate learning rates, training becomes highly unstable, and without identity initialization, training is less efficient.}
\label{fig:koopman_ablation}
\end{figure*}


We use the flow-based \textit{K-UBM} and present an ablation study across simulation tasks and four real-world tasks by training models using the same best-performing random seed. In Fig.~\ref{fig:koopman_ablation}, \textit{Identity init + separate LR} denotes the training configuration used throughout the main paper, which incorporates both key factors. \textit{Separate LR only} corresponds to using separate learning rates for the lifting network and Koopman matrix without identity initialization. \textit{Identity init only} initializes the Koopman matrix as the identity but uses a shared learning rate, while \textit{Neither} disables both.

From the results, we first observe that without separate learning rates, training becomes highly unstable across all 11 tasks (7 sim and 4 real), as evidenced by frequent spikes in both the training loss and the spectral radius of the Koopman matrix. 

Next, when only separate learning rates are used without identity initialization, the training remains stable but is significantly less efficient (note that the training loss is plotted on a logarithmic scale), and we observe a sharp increase of the spectral radius from its initial value early in training. These results indicate that stable and accurate long-horizon prediction requires the Koopman matrix to maintain sufficient expressivity while avoiding excessive energy dissipation during rollout. In practice, this corresponds to keeping the spectral radius close to one~\cite{mezic2020koopman}, which is effectively achieved by initializing the Koopman matrix as the identity.
\section{Replanning Trigger Detection Results}
\label{sec:Replan_trigger_detection}
\begin{figure}[t]
\centering
\includegraphics[width=0.9\textwidth]{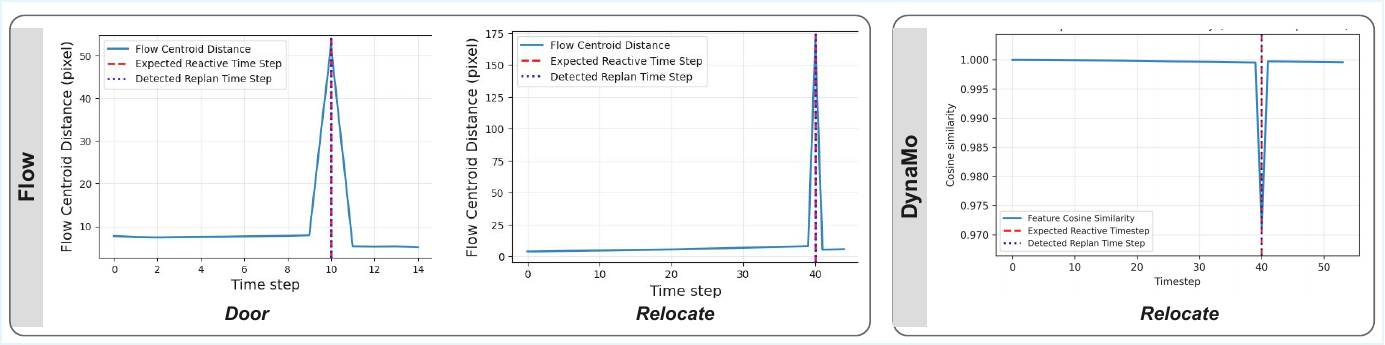}
\caption{For both visual features, we demonstrate how replanning triggers can be detected. Left: For the flow-based \textit{K-UBM} policy, we show a sharp change in flow centroid distance. Right: Similarly, for DynaMo features, we visualize the corresponding changes in cosine similarity.}
\label{fig:flow_replan_trigger}
\end{figure}

As described in Section~\ref{appendix:Reactivity}, we can observe the sharp discrepancies between the predicted visual states and the actual visual observations when the environment changes. The results in Fig.~\ref{fig:flow_replan_trigger} illustrate such sharp changes that K-UBM can precisely detect as replanning triggers, enabled by its accurate object motion prediction.

Specifically, we monitor the flow-space divergence
$d_t = \lVert \mathrm{centroid}(\text{observed flow}) -
\mathrm{centroid}(\text{predicted flow}) \rVert$ at every step, and trigger
replanning when the latest jump $|d_t - d_{t-1}|$ exceeds $3\times$ the running
mean of the previous jumps. Since the rule is normalized by each episode's own
divergence history, there is no absolute pixel threshold to tune.

\section{Robustness to sensory degradation}
\label{sec:Perturbation}
In this section, we conduct experiments to compare reactive policies with K-UBM policies in terms of robustness to sensory degradation. The key distinction between the two paradigms lies in the trade-off between \textit{robustness} and \textit{reactivity}. Reactive policies rely on real-time sensory feedback at a fixed control frequency, which increases the likelihood of observing inaccurate sensory inputs, while enabling adaptation to environmental changes. In contrast, K-UBM policies support open-loop execution and can thus remain agnostic to online sensor noise, but typically at the cost of reactivity. However, as described in the Section.~\ref{appendix:Reactivity}, the ability to predict environment dynamics combined with our replanning mechanism endows K-UBM with \textit{framework-level} reactivity. 

Specifically, we evaluate the robustness of diffusion- and ACT-based policies across all simulation tasks under increasing levels of visual corruption, modeled as blacked-out images to mimic potential camera failures in real-world manipulation scenarios. For each task, visual feature, and model, we select the best-performing architecture and seed. We omit analogous experiments for K-UBM, as it supports open-loop rollout and does not strictly require online feedback during execution.

\begin{figure*}[t]
\centering
\includegraphics[width=\textwidth]{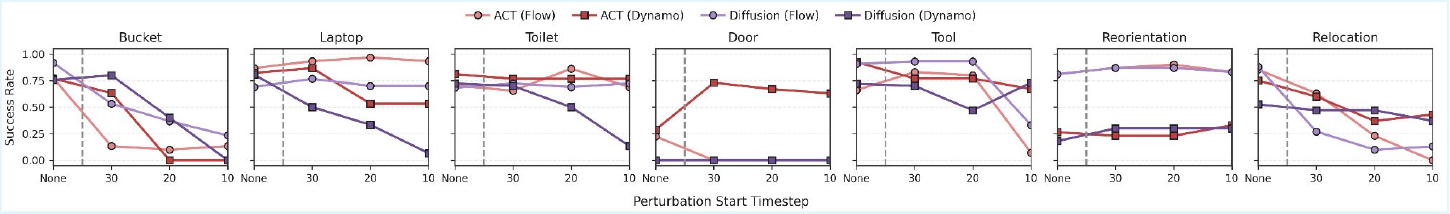}
\caption{For both visual features, we observe that performance of reactive policies often drops significantly under corrupted visual inputs, with earlier corruption leading to more severe degradation. The only exception is the Door task, where performance improves under noise. However, we find this is due to the policy exploiting unrealistic behaviors in simulation.}
\label{fig:perturbation}
\end{figure*}

As shown in Fig.~\ref{fig:perturbation}, the performance of both policies drops significantly across all tasks under corrupted visual inputs. This strongly supports a key advantage of the UBM frameworks: robustness to inaccurate online feedback compared to reactive policies.


\section{Real-world Per-task Result}
\label{appendix:real-world-per-task}
We report the real-world per-task results in Table.~\ref{tab:realworld_pertask}.
\begin{table*}[h]
\centering
\caption{\textbf{Per-task real-world results.} Success rate (\%, $\uparrow$),
arm and hand RMS jerk (rad/s$^3$, $\downarrow$). Jerk cells are mean$\pm$std over
10 runs; \textbf{bold}/\underline{underline} = best/second-best per task.}
\label{tab:realworld_pertask}
\begingroup
\small
\setlength{\tabcolsep}{5pt}
\renewcommand{\arraystretch}{1.15}
\providecommand{\mci}[2]{#1{\tiny\textcolor{black!55}{$\pm$#2}}}
\renewcommand{\mci}[2]{#1{\tiny\textcolor{black!55}{$\pm$#2}}}
\resizebox{0.92\textwidth}{!}{%
\begin{tabular}{@{}ll ccccc@{}}
\toprule
\textbf{Metric} & \textbf{Method} & \textbf{Flower} & \textbf{Chips} & \textbf{Whiteboard} & \textbf{Dinner} & \textbf{Avg} \\
\midrule
\multirow{3}{*}{\textbf{Success $\uparrow$}}
 & Diffusion (2,8,16) & 70.0 & \textbf{100.0} & 40.0 & \underline{80.0} & 72.5 \\
 & ACT (40)           & \textbf{100.0} & \underline{90.0} & \underline{80.0} & \textbf{100.0} & \textbf{92.5} \\
 & K-UBM (Ours)       & \underline{90.0} & 70.0 & \textbf{100.0} & \underline{80.0} & \underline{85.0} \\
\midrule
\multirow{3}{*}{\textbf{Arm Jerk $\downarrow$}}
 & Diffusion (2,8,16) & \mci{\underline{27.33}}{17.32} & \mci{44.00}{8.55} & \mci{77.57}{21.79} & \mci{29.88}{6.42} & \mci{44.69}{23.11} \\
 & ACT (40)           & \mci{\underline{27.33}}{7.14}  & \mci{\underline{22.15}}{7.38} & \mci{\underline{40.19}}{14.55} & \mci{\underline{24.00}}{5.03} & \mci{\underline{28.42}}{8.14} \\
 & K-UBM (Ours)       & \mci{\textbf{5.55}}{0.60} & \mci{\textbf{7.42}}{1.34} & \mci{\textbf{4.30}}{0.26} & \mci{\textbf{2.65}}{0.24} & \mci{\textbf{4.98}}{2.01} \\
\midrule
\multirow{3}{*}{\textbf{Hand Jerk $\downarrow$}}
 & Diffusion (2,8,16) & \mci{\underline{12.03}}{6.84} & \mci{8.57}{2.64} & \mci{16.51}{5.58} & \mci{\underline{19.55}}{8.04} & \mci{\underline{14.16}}{4.84} \\
 & ACT (40)           & \mci{13.56}{12.69} & \mci{\underline{5.43}}{3.22} & \mci{\underline{12.68}}{9.99} & \mci{34.59}{10.72} & \mci{16.56}{12.56} \\
 & K-UBM (Ours)       & \mci{\textbf{3.84}}{0.33} & \mci{\textbf{2.88}}{0.63} & \mci{\textbf{3.21}}{0.36} & \mci{\textbf{6.65}}{0.94} & \mci{\textbf{4.14}}{1.71} \\
\bottomrule
\end{tabular}}
\endgroup
\end{table*}
\end{document}